# A Integrating CRITIC-WASPAS group decision-making method under interval-valued q-Rung orthogonal fuzzy enviroment

**Benting Wan, shufen Zhou**

School of software and Internet of things Engineering, Jiangxi University of Finance and Economics, Nanchang 330013, China;

Correspondence should be addressed to Benting Wan: wanbenting@jxufe.edu.cn

**Abstract**: This paper provides a new tool for multi-attribute multi-objective group decision-making with unknown weights and attributes' weights. An interval-valued generalized orthogonal fuzzy group decision-making method is proposed based on the Yager operator and CRITIC-WASPAS method with unknown weights. The method integrates Yager operator, CRITIC, WASPAS, and interval value generalized orthogonal fuzzy group. Its merits lie in allowing decision-makers greater freedom, avoiding bias due to decision-makers' weight, and yielding accurate evaluation. The research includes: expanding the interval value generalized distance measurement method for comparison and application of similarity measurement and decision-making methods; developing a new scoring function for comparing the size of interval value generalized orthogonal fuzzy numbers, and further existing researches. The proposed interval-valued Yager weighted average operator (IVq-ROFYWA) and Yager weighted geometric average operator (IVq-ROFYWG) are used for information aggregation. The CRITIC-WASPAS combines the advantages of CRITIC and WASPAS, which not only work in the single decision but also serve as the basis of the group decision. The in-depth study of the decision-maker's weight matrix overcomes the shortcomings of taking the decision as a whole, and weighs the decision-maker's information aggregation. Finally, the group decision algorithm is used for hypertension risk management. The results are consistent with decision-makers' opinions. Practice and case analysis have proved the effectiveness of the method proposed in this paper. At the same time, it is compared with other operators and decision-making methods, which proves the method effective and feasible.

## 1. Introduction

Decision-making (DM) is widespread in human production and life[49-50]. Decision-making problems include multiple alternatives, each of which contains various attributes. The MADM problem, involving multiple people, is called the multi-attribute multi-objective group decision-making (MAGDM) problem [55-56]. Since Zadeh proposed fuzzy sets [1], fuzzy theories and methods have been widely used in MADM and MAGDM[53-54]. In order to express the decision makers' approval, disapproval, and hesitation, Atanassov proposed the Intuitionistic Fuzzy Set (IFS)[2], where approval and disapproval are expressed by membership degree （$u \geq 0$） and non-membership degree （$v \geq 0$） respectively. Moreover $u + v \leq 1$, it provides more freedom for group decision-making applications. For example, Kamal Kumar and Chen[6] studied Einstein's MADM method based on intuitionistic fuzzy values, Yager et al.[11] studied MADM method of generalized intuitionistic fuzzy sets. Liu et al. [10] also proposed a group decision-making method based on the Heronian operator of intuitionistic fuzzy numbers. Seda Türk et al.[47] used intuitionistic fuzzy sets for the MAGDM problem with multiple photovoltaic locations. Yager proposed the Pythagorean fuzzy set [8-9], $u^2 + v^2 \leq 1$, and proposed generalized orthogonal fuzzy sets in 2017[12], $u^q + v^q \leq 1$. Generalized orthogonal fuzzy sets not only integrate intuitionistic fuzzy sets and Pythagorean fuzzy sets, but also expand the application field and provide decision-makers with more free representation [39-40]. Although the intuitionistic fuzzy set can express the decision-maker's approval, disapproval, and hesitation, when the decision-maker is not sure, it is often represented by an interval value. For this reason, the researchers proposed the interval value generalized orthogonal fuzzy set [46], in MAGDM Group decision-making has been well applied. Jie Wang et al.[41] proposed a MAGDM method based on interval-valued generalized orthogonal fuzzy, and applied it to the selection of green suppliers. Jin et al.[32] proposed interval-valued generalized orthogonality The application of fuzzy in the risk assessment of improved tool change manipulators, Limei Liu et al.[33] proposed a large-scale green supplier selection method in an interval-value generalized orthogonal fuzzy environment. The conditions of these group decision-making methods are that the decision-maker's weight and attribute weight are known. There is no interval-valued generalized orthogonal fuzzy group decision-making method with unknown decision-maker's weight and unknown attribute weight. This paper will make up for this deficiency.

In MAGMD group decision-making, operators are often used to fuse or aggregate the information of decision-makers. Researchers propose some classic aggregators, such as OWA[60], OWG[61], Frank[58], Einsten[59], Heronian[63], Bonferroni [62], Choquet[64], Yager[22] and other operators. In different fuzzy sets, researchers, such as Fei and Feng[57], took a step further and used Dempster Rule aggregation to solve intuitionistic fuzzy value problem and proposed a MOS operator based on OWA operator. Wei et al.[51] proposed a hesitant Pythagorean fuzzy interaction operator based on Bonferroni operator: HPFIWB operator and HPFIGWBM operator , And applied to the MAGDM problem. Chiranjibe Jana et al. [52] proposed a Dombi-based Pythagorean aggregator and its application in multi-attribute decision making. Liu et al. [43] proposed a regular swing hesitation fuzzy language weighted mean aggregation operator based on the Hamy operator and its application in multi-attribute decision-making and defined NWHFLPHAM operator and NWHFLPWHAM operator. Liu et al.[42] studied generalized orthogonal Maclaurin symmetric mean operators: q-ROFWGMSM operator, q-ROFWGGMSM operator, and they were applied in MAGDM. Jin et al. [32] proposed the interval-valued generalized orthogonal IVq-ROFWMSM operator, combined it with the interval-valued generalized orthogonal IVq-ROFWG operator to fuse the evaluation information of decision-makers, and combined it with the decision-making method to obtain the result of the scheme ranking. Jie Wang et al. [41] proposed interval-valued generalized orthogonal q-RIVOFWA operators, interval-valued generalized orthogonal q-RIVOYWG operators, interval-valued generalized orthogonal q-RIVOFHM operators, interval-valued generalized orthogonal q-RIVOFHM operators, and interval-valued generalized orthogonal q-RIVOFHM operators based on Hamy operator. He also put forward the q-RIVOFWHM operator, interval-valued generalized orthogonal q-RIVOFDHM operator and interval-valued generalized orthogonal q-RIVOFWDHM operator, and then used q-RIVOFWA operator to fuse information for decision. According to the fusion information, the sub-programs can be sorted. Among the many operators, the Yager operator is also a very effective aggregation technology, which can control the uncertainty of some decision-makers' evaluation data, so as to obtain dense information. The aggregation ability of the Yager operator can eliminate the uncertainty of the decision makers' evaluation data, and aggregate the decision results of multiple decision makers in a certain proportion. Therefore, some scholars have applied it to fuzzy sets: Liu et al.[14] proposed q-RPFYWA operators, q-RPFYOWA operators, q-RPFYWG operators, q-RPFYWG operators, and q-RPFYWG operators based on generalized orthogonal fuzzy numbers based on Yager operations. RPFYHWG operator, and deal with multi-attribute decision-making problems in a modern way. Muhammad Akram and Peng et al. [44] studied the multi-criteria decision-making model based on the Pythagorean fuzzy Yager aggregation operator, but no scholars have proposed interval-valued generalized orthogonal fuzzy operators based on the Yager algorithm.

Although operators can be used for information fusion, attribute weights are still required in the assembly process. In addition, when attributes have special relationships, operators cannot meet the needs of decision-making. Therefore, when the decision-making environment is relatively complex, corresponding methods are often used. At present, researchers have proposed a large number of decision-making methods, such as TOPSIS method[24], MOORA method[18], COPRAS method[27], MACBETH method[45], The MABAC method[17], et al. Among them, the TOPSIS determines the method priority according to the information of the decision matrix and the distance between the positive ideal (PIS) and the negative ideal (NIS). TOPSIS has been extended to the Pythagorean fuzzy set[3], intuitionistic fuzzy sets[38], and other fuzzy sets. The MOORA decision-making method is a multi-objective optimization based on proportional analysis. There are corresponding studies in the fuzzy field: Hasan et al.[69] proposed a type 2 hybrid fuzzy evaluation of the financial service interval of the E7 economy based on the DEMATEL-ANP and MOORA methods. The COPRAS decision-making method considers the impact of maximizing and minimizing the attribute index on the result evaluation and sorts the schemes according to the magnitude of the significant value. There have been a lot of research results: for example, Pratibha Rani et al. proposed a new COPRAS method to solve the problem. For the problem of drug selection for type 2 diabetes under the Grass fuzzy environment[67], Adjei Peter Darko and Decui Liang proposed the COPRAS method of dual hesitation and fuzzy[68]. The MACBETH decision method optimizes data normalization by defining the Macbeth score and determining the order of the schemes through the Macbeth score. The MABAC decision-making method ranks and selects the best for each plan based on the difference between the distances. Gong et al. proposed the review and evaluation of undergraduate teaching based on the extended MABAC method in a generalized orthogonal fuzzy environment[65], and Jia et al. proposed a method based on intuitionistic fuzzy of the MABAC with extended multi-criteria group decision-making method[66]. The CRITIC method[15] and WASPAS method[16] in many decision-making methods are also very useful. The CRITIC method can determine the weight of each attribute according to the relationship between the attributes. This method can eliminate some highly correlated ones. The impact of indicators, reducing the overlap of information between indicators, is conducive to obtaining credible evaluation results, but the evaluation of the program cannot be determined. The WASPAS method is a combination of the weighted sum model (WSM) and the weighted product model (WPM). It is the latest multi-objective optimization method with higher accuracy. However, the overall evaluation of each alternative can be determined after the weight is known. Therefore, the CRITIC method and the WASPAS method are combined to form a new CRITIC-WASPAS method, which can calculate the weight between attributes according to the decision matrix given by the decision maker, and then make the best choice of alternatives. At present, many researchers have carried out research on fuzzy sets for these two methods. For example, to solve the multi-criteria decision-making on the Pythagorean fuzzy set, the CRITIC method is used to calculate the objective weight[35], and the interval value fuzzy soft decision intelligent medical management evaluation method based on CoCoSo and CRITIC is proposed[36]. Peng[13] used the CRITIC method to calculate objective weights in financial risk assessment decisions on generalized orthogonal fuzzy sets. Katarzyna Rudnik et al.[37] proposed the OFN-WASPAS method, that is the WASPAS method using ordered fuzzy numbers. R. Davoudabadi et al. [38] developed a new method of aggregation and ranking based on WASPAS and TOPSIS methods and applied it on interval-valued intuitionistic fuzzy sets. However, no scholar has studied the application of CRITIC method and WASPAS method in interval-valued generalized orthogonal fuzzy sets.

Measurement methods such as distance, entropy, and similarity are used to compare and analyze fuzzy numbers. Researchers have conducted a lot of research. Among them, Peng and Liu [4] summarized the generalized orthogonal measurement method, and carried out the generalized orthogonal distance measurement method. In-depth analysis, according to Topsis theory and reference point theory, researchers have proposed positive ideal distance (PIS), negative ideal distance (NIS), and center-selected distance (CIS) [24]. As a result, the negative ideal distance examines the result of people and total opposition, while the middle distance uses parameters to establish the relationship between the positive ideal distance and the negative one. The selected parameters need to be judged according to risk preference, and more decisions are required. Judgment increases the cost of decision-making. Therefore, the ideal distance measurement method is often used to determine the weight of decision makers. Yue [70] proposed a method based on the ideal distance to obtain the similarity for the decision makers' weights. He also [71] proposed a mapping method to solve the decision makers' weights. The methods described in documents [70] and [71] are based on individual decisions. The decision maker's evaluation of the overall plan to solve the decision maker's weight is widely used in MAGDM problem processing. However, this method does not consider the difference of decision makers in evaluating different plans. In order to meet the decision maker's demand for deviations in different plans, this article has developed the method of acquiring the weight of decision makers based on the similarity of the ideal distances makes up for the shortcomings of the existing methods.

At the same time, in order to be able to compare the size of fuzzy numbers, the researchers introduced a scoring function. Many scholars have yielded their own results in this regard. Li et al[5] put forward an information-based IVIFS score function by comprehensively considering the amount of information, reliability, deterministic information and relative closeness. Chen [28] overcome the shortcomings of the existing score function and proposed a new IVIFS one. At present, there are many score functions for interval-valued fuzzy sets, but some of them inevitably have shortcomings. For example, the score function proposed by Cheng [29], Bai [30], and Gong and Ma [31] can not compare fuzzy numbers for certain interval values. To overcome this problem, this paper proposes a new interval-valued fuzzy set scoring function.

At present, the existing group decision-making methods include researches based on operators, decision methods, and the combination of both. Sometimes, the weight of the decision maker is unknown, the attribute weight is unknown, and the weight of the decision maker related to the attribute weights are all unknown, but no scholars have proposed a group decision-making method for the unknown weights and attribute weights under interval-valued generalized orthogonal fuzzy sets. Therefore, this paper proposes an interval-valued generalized orthogonal fuzzy group decision-making method that integrates the Yager aggregation operator and the CRITIC-WASPAS method. In the decision-making process, dense data information can be obtained, credible attribute weights and higher-precision program evaluations can be obtained. The research contributions of this paper

are as follows:

(1) Inspired by the literature [4], a distance measurement method based on interval-valued generalized orthogonal fuzzy has been developed. The measurement methods of the positive ideal distance, negative ideal distance, and compromise distance of the number are the basis for solving the weight of decision makers in this paper, which are described in Section 3.1 of this paper.

(2) Inspired by the literature [28], this paper proposes a new interval-valued fuzzy number score function. The proposed score function not only makes up for the shortcomings of [29], [30], [31], and works very well. To compare the size of two interval-valued fuzzy numbers, it can be well used in the ranking of alternatives, which is explained in section 3.2 of this article.

(3) Inspired by the literature [70], this paper proposes a method of weighting decision makers based on the similarity of different plans based on ideal distance, and derives a weight matrix of decision makers' plans. The proposed decision-maker weight method not only avoids the lack of equal weights in the overall solution, but also reflects the differences in the evaluation of different options by decision-makers. When the information of different decision-makers is assembled, the decision-maker's plan weight matrix is used, which has a good application uwhen the decision-makers have bias in evaluating different objects. This method will be described in section 6.2 of this article.

(4) Inspired by the literature [14], this paper proposes the Yager algorithm and algorithm of interval-valued generalized orthogonal fuzzy numbers, took a step further to raise the Yager weighted average operator of interval-valued generalized orthogonal fuzzy numbers ( IVq-ROFYWA) and the weighted geometric average operator (IVq-ROFYWG), and illustrate the properties of the two operators in detail. The IVq-ROFYWA and IVq-ROFYWG operators proposed in this paper compensate for the generalized orthogonality of the Yager operator in the interval value. The blanks in fuzzy numbers are explained in Section 4 of this article.

(5) Inspired by the literature [15] and [16], this paper proposes a CRTIC-WASPAS decision-making method based on interval-valued generalized orthogonal fuzzy numbers. The CRTIC-WASPAS fusion method studied not only combines the CRITIC and WASPAS methods, but also incorporates IVq-ROFYWA and IVq-ROFYWG operators into the decision-making process. The CRTIC-WASPAS fusion method uses the CRITIC method to solve the interval value weight of the attribute, and then uses the ideal distance to convert it into a real number weight, and finally uses WASPAS to rank alternatives according to the real number weight of the attribute. Therefore, it effectively solves the MADM problem of unknown attribute weights, which is described in Section 5 of this article.

(6) This paper proposes an interval-valued generalized orthogonal fuzzy group decision-making method integrating IVq-ROFYWA, IVq-ROFYWG operator and CRITIC-WASPAS fusion method. Using the proposed method for solving decision makers' weights and the IVq-ROFYWA operator to aggregate the decision information of multiple decision makers, the aggregated information is obtained through the CRITIC-WASPAS fusion method to obtain interval-valued generalized orthogonal fuzzy numbers for different schemes, and finally used in this paper. The scoring function ranks the evaluation of the schemes and obtains the optimal scheme. This group decision-making method only requires the decision maker to provide a decision matrix based on the problem, and does not require any further information on weight or attribute weight information. Therefore, it greatly improves the efficiency of decision-making as it can be used to solve the unknown weight of the decision maker and the attribute. The interval-valued generalized orthogonal MAGDM problem with unknown weights is described in Section 6 of this paper.

(7) Apply the proposed group decision-making method to the early warning management of hypertension risk diseases. According to the decision-makers based on the patient's inspection results and risk factors, they will give a decision matrix and their own judgments. The group decision-making method proposed in this paper is used to calculate the risk level. The result is consistent with the decision maker's prediction, which proves the effectiveness of the method proposed in this paper. At the same time, the plan is analyzed in detail. The experimental results confirm that the group decision method proposed in this paper is highly adaptable, and it is compared and analyzed with the existing operators and decision methods. It further proves that the decision-making method proposed in this paper is effective.

This article is organized as follows: Section 2 explains the basic knowledge needed for this research, including interval-valued generalized orthogonal arithmetic and Yager operator. Section 3 describes the proposed interval-valued fuzzy score function and distance measurement method. Section 4 describes the interval-valued generalized orthogonal Yager operator. Section 5 describes the interval-valued generalized orthogonal CRITIC method and the WASPAS method, and combines both advantages to propose the CRITIC-WASPAS decision method. Chapter 6 describes the integrated group decision-making method. Section 7 expounds the application of the group decision-making method in hypertension management, and analyzes and verifies the group decision-making method proposed in this paper.

## 2. Preliminary knowledge

### 2.1 Interval-valued generalized orthogonal fuzzy sets

**Definition 2.1** [19]: Let X be the universe of discourse, and $A = \{\langle x, u_A(x), v_A(x)\rangle | x \in X\}$ be the generalized orthogonal fuzzy set （$q - ROFS$） on the domain X, where $u_A$: $X \to [0,1]$, $v_A$: $X \to [0,1]$ is the two mappings on X and satisfies formula (1).

$$0 \leq (u_A(x))^q + (v_A(x))^q \leq 1 \tag{1}$$

In formula (1), $q \geq 1$, $u_A(x)$ is the degree of membership of $X \to [0,1]$; $v_A(x)$ is the degree of non-membership of $X \to [0,1]$, and the degree of hesitation $\pi_A(x)$, as shown in formula (2):

$$\pi_A(x) = \sqrt[q]{1 - (u_A(x))^q + (v_A(x))^q} \quad (q \geq 1) \tag{2}$$

**Definition 2.2** [20]: Let X be the universe of discourse, and the interval-valued generalized orthogonal fuzzy set （$IVq - ROFS$） A on X is defined as shown in formula (3).

$$A = \{<x, u_A(x), v_A(x)> | x \in X\} \tag{3}$$

In formula (3), the membership function is a mapping of interval values that satisfies: $u_A(x) = [u_A^-(x), u_A^+(x)] \subseteq [0,1]$, and the non-membership function is a mapping of interval values that satisfies: $v_A(x) = [v_A^-(x), v_A^+(x)] \subseteq [0,1]$, and at the same time satisfies: $0 \leq (u_A^+(x))^q + (v_A^+(x))^q \leq 1, (q \geq 1)$。 The hesitation of set A is shown in formula (4).

$$\pi_A(x) = [\pi_A^-(x), \pi_A^+(x)] = [\sqrt[q]{1 - (u_A^+(x))^q - (v_A^+(x))^q}, \sqrt[q]{1 - (u_A^-(x))^q - (v_A^-(x))^q}] \tag{4}$$

**Definition 2.3** [21]: Let $a = ([u^-, u^+], [v^-, v^+])$、 $a_1 = ([u_{a_1}^-, u_{a_1}^+], [v_{a_1}^-, v_{a_1}^+])$ and $a_2 = ([u_{a_2}^-, u_{a_2}^+], [v_{a_2}^-, v_{a_2}^+])$ be three $IVq - ROFNs$, where $q \geq 1$，then the interval-valued generalized orthogonal fuzzy set addition, multiplication, number multiplication, power operation rules are as

(5), (6), ( 7) As shown in (8).

$$a_1 \oplus a_2 = \left( \begin{array}{c} [\sqrt[q]{(u_{a_1}^-)^q + (u_{a_2}^-)^q - (u_{a_1}^-)^q(u_{a_2}^-)^q}, \sqrt[q]{(u_{a_1}^+)^q + (u_{a_2}^+)^q - (u_{a_1}^+)^q(u_{a_2}^+)^q}], \\ [v_{a_1}^- v_{a_2}^-, v_{a_1}^+ v_{a_2}^+] \end{array} \right) \quad (5)$$

$$a_1 \otimes a_2 = \left( \begin{array}{c} [u_{a_1}^- u_{a_2}^-, u_{a_1}^+ u_{a_2}^+], \\ [\sqrt[q]{(v_{a_1}^-)^q + (v_{a_2}^-)^q - (v_{a_1}^-)^q(v_{a_2}^-)^q}, \sqrt[q]{(v_{a_1}^+)^q + (v_{a_2}^+)^q - (v_{a_1}^+)^q(v_{a_2}^+)^q}] \end{array} \right) \quad (6)$$

$$\lambda a = <[\sqrt[q]{1-(1-(u^-)^q)^\lambda}, \sqrt[q]{1-(1-(u^+)^q)^\lambda}], [(v^-)^\lambda, (v^+)^\lambda]> \quad (7)$$

$$a^\lambda = <[(u^-)^\lambda, (u^+)^\lambda], [\sqrt[q]{1-(1-(v^-)^q)^\lambda}, \sqrt[q]{1-(1-(v^+)^q)^\lambda}]> \quad (8)$$

Inspired by the subtraction and division operations of generalized orthogonal fuzzy sets defined by Peng [13], this paper proposes interval-valued generalized orthogonal subtraction and division, as shown in Definition 2.4.

**Definition 2.4:** Let $a_1 = [u_{a_1}^-, u_{a_1}^+], [v_{a_1}^-, v_{a_1}^+]$, $a_2 = [u_{a_2}^-, u_{a_2}^+], [v_{a_2}^-, v_{a_2}^+]$ be 2 IVq-ROFNs, and $q \geq 1$, the subtraction and division rules of IVq-ROFNs are shown in formulas (9) and (10).

$$a_1 \ominus a_2 = \left( \frac{[(u_{a_1}^-)(v_{a_2}^-),(u_{a_1}^+)(v_{a_2}^+)],}{\sqrt[q]{(v_{a_1}^+)^q + (u_{a_2}^+)^q - ((v_{a_1}^+)^q(u_{a_2}^+)^q)}}, \sqrt[q]{(v_{a_1}^-)^q + (u_{a_2}^-)^q - ((v_{a_1}^-)^q(u_{a_2}^-)^q)} \right)$$

$$a_1 \oslash a_2 = \left( \sqrt[q]{(u_{a_1}^+)^q + (v_{a_2}^+)^q - ((u_{a_1}^+)^q(v_{a_2}^+)^q)}], \frac{[\sqrt[q]{(u_{a_1}^-)^q + (v_{a_2}^-)^q - ((u_{a_1}^-)^q(v_{a_2}^-)^q)},}{[(v_{a_1}^-)(u_{a_2}^-),(v_{a_1}^+)(u_{a_2}^+)]} \right)$$

(10)

**Theorem 1:** The calculation results of formulas (9) and (10) are still interval-valued generalized orthogonal fuzzy numbers.

**Proof:** According to formula (9), set $a_1 \ominus a_2 = <[u^-, u^+], [v^-, v^+]>$, we can get:

$$(u^-)^q + (v^-)^q = ((u_{a_1}^-)(v_{a_2}^-))^q + \left( \sqrt[q]{(v_{a_1}^-)^q + (u_{a_2}^-)^q - ((v_{a_1}^-)^q(u_{a_2}^-)^q)} \right)^q$$

$$= (u_{a_1}^-)^q(v_{a_2}^-)^q + (v_{a_1}^-)^q + (u_{a_2}^-)^q - (v_{a_1}^-)^q(u_{a_2}^-)^q$$

$$= (u_{a_1}^-)^q(v_{a_2}^-)^q + (u_{a_2}^-)^q + (v_{a_1}^-)^q(1 - (u_{a_2}^-)^q) \geq 0$$

$$(u^-)^q + (v^-)^q = (u_{a_1}^-)^q(v_{a_2}^-)^q + (u_{a_2}^-)^q + (v_{a_1}^-)^q(1 - (u_{a_2}^-)^q)$$

$$= (u_{a_1}^-)^q(v_{a_2}^-)^q + (u_{a_2}^-)^q + (v_{a_1}^-)^q(1 - (u_{a_2}^-)^q)$$

$$\leq (u_{a_1}^{-})^q \big(1 - (u_{a_2}^{-})^q\big) + (u_{a_2}^{-})^q + (v_{a_1}^{-})^q \big(1 - (u_{a_2}^{-})^q\big)$$

$$\leq \big((u_{a_1}^{-})^q + (v_{a_1}^{-})^q\big)\big(1 - (u_{a_2}^{-})^q\big) + (u_{a_2}^{-})^q$$

$$\leq 1 * \big(1 - (u_{a_2}^{-})^q\big) + (u_{a_2}^{-})^q = 1$$

It can be found $0 \leq (u^-)^q + (v^-)^q \leq 1$, which shows that the result of formula (9) is interval-valued generalized orthogonal fuzzy number. Similarly, it can be proved that the calculation result of formula (10) is also interval-valued generalized orthogonal fuzzy number. The theorem is proved.

## 2.2 Yager operator

The Yager operator can control the uncertainty of some decision makers' evaluation data, thereby obtaining dense information. Under the environment of dense information, it is used for the aggregation of decision makers' information for MAGDM problems.

**Definition 2.5[22]:** For any two real numbers a,b ∈ [0,1],Yager sum $\oplus_Y$ and Yager product $\otimes_Y$,The expression of is t-norms operation and t-conorms operation, as shown in (11) and (12).

$$a \oplus_Y b = \min\left[1, (a^p + b^p)^{\frac{1}{p}}\right], p \in [1, +\infty] \quad (11)$$

$$a \otimes_Y b = 1 - \min\left\{1, [(1-a)^p + (1-b)^p]^{\frac{1}{p}}\right\}, p \in [1, +\infty] \quad (12)$$

Yager confirmed that when $p = 1$, ($\oplus_Y, \otimes_Y$) will be transformed into ($\oplus, \otimes$), that is, t-norms operation and t-conorms operation are performed on $a$ and $b$,When $p \to +\infty$, ($\oplus_Y, \otimes_Y$) will be transformed into (∨,∧), that is, take the larger and smaller operations of $a$ and $b$

## 3. Interval distance and score function

### 3.1 Interval value generalized orthogonal fuzzy distance

Distance is used to measure the degree of difference between two generalized orthogonal fuzzy sets in the theory of generalized orthogonal fuzzy sets. The weights of decision-makers and decision-making methods in this paper need to solve the distance between IVq-ROFNs. Inspired by the D2 distance formula given in Peng[4],this paper proposes a new interval-valued generalized orthogonal distance definition. The formula definition is shown in 3.1.

**Definition 3.1:** For any two interval values of $a_1 = <[u_{a_1}^-, u_{a_1}^+], [v_{a_1}^-, v_{a_1}^+]>$ and $a_2 = <[u_{a_2}^-, u_{a_2}^+], [v_{a_2}^-, v_{a_2}^+]>$, the generalized orthogonal fuzzy number, where $q \geq 1$, the distance is shown in formula (13).

$$d(a_1, a_2) = \frac{1}{4}(\big|(u_{a_1}^-)^q - (u_{a_2}^-)^q - ((v_{a_1}^-)^q - (v_{a_2}^-)^q)\big| +$$

$$\left|(u_{a_1}{}^+)^q - (u_{a_2}{}^+)^q - \left((v_{a_1}{}^+)^q - (v_{a_2}{}^+)^q\right)\right|) \quad (13)$$

**Theorem 2:** Formula (13) satisfies the following properties:

（1） $0 \leq d(a_1, a_2) \leq 1$;

（2） $d(a_1, a_2) = d(a_2, a_1)$;

（3） $d(a_1, a_2) = 0$, only if $a_1 = a_2$;

（4） If $a_1 \subseteq a_2 \subseteq a_3$, then $d(a_1, a_2) \leq d(a_1, a_3)$ and $d(a_2, a_3) \leq d(a_1, a_3)$.

Since the nature (2) and nature (3) of distance are easy to prove, this article only gives the proof process of nature (1) and nature (4).

**Proof:**

1）：$d(a_1, a_2) = \dfrac{1}{4}\left\{\begin{array}{l}\left|(u_{a_1}{}^-)^q - (u_{a_2}{}^-)^q - \left[(v_{a_1}{}^-)^q - (v_{a_2}{}^-)^q\right]\right| + \\ \left|(u_{a_1}{}^+)^q - (u_{a_2}{}^+)^q - \left[(v_{a_1}{}^+)^q - (v_{a_2}{}^+)^q\right]\right|\end{array}\right\}$

$= \dfrac{1}{4}\left[\begin{array}{l}\left|(u_{a_1}{}^-)^q - (v_{a_1}{}^-)^q + (v_{a_2}{}^-)^q - (u_{a_2}{}^-)^q\right| + \\ \left|(u_{a_1}{}^+)^q - (v_{a_1}{}^+)^q + (v_{a_2}{}^+)^q - (u_{a_2}{}^+)^q\right|\end{array}\right]$

$\leq \dfrac{1}{4}\left\{\begin{array}{l}\left|\max\left[(u_{a_1}{}^-)^q - (v_{a_1}{}^-)^q\right] + \max\left[(v_{a_2}{}^-)^q - (u_{a_2}{}^-)^q\right]\right| + \\ \left|\max\left[(u_{a_1}{}^+)^q - (v_{a_1}{}^+)^q\right] + \max\left[(v_{a_2}{}^+)^q - (u_{a_2}{}^+)^q\right]\right|\end{array}\right\}$

$\leq \dfrac{1}{4} * (2 + 2) = 1$

When the two fuzzy numbers are exactly the same, and the distance is obviously 0, so: $0 \leq d(a_1, a_2) \leq 1$, Proven.

（4）：$d(a_1, a_2) = \dfrac{1}{4}\left\{\begin{array}{l}\left|(u_{a_1}{}^-)^q - (u_{a_2}{}^-)^q - \left[(v_{a_1}{}^-)^q - (v_{a_2}{}^-)^q\right]\right| + \\ \left|(u_{a_1}{}^+)^q - (u_{a_2}{}^+)^q - \left[(v_{a_1}{}^+)^q - (v_{a_2}{}^+)^q\right]\right|\end{array}\right\}$

$= \dfrac{1}{4}\left\{\begin{array}{l}(v_{a_1}{}^-)^q - (v_{a_2}{}^-)^q - \left[(u_{a_1}{}^-)^q - (u_{a_2}{}^-)^q\right] + \\ (v_{a_1}{}^+)^q - (v_{a_2}{}^+)^q - \left[(u_{a_1}{}^+)^q - (u_{a_2}{}^+)^q\right]\end{array}\right\}$

$d(a_1, a_3) = \dfrac{1}{4}\left\{\begin{array}{l}\left|(u_{a_1}{}^-)^q - (u_{a_3}{}^-)^q - \left[(v_{a_1}{}^-)^q - (v_{a_3}{}^-)^q\right]\right| + \\ \left|(u_{a_1}{}^+)^q - (u_{a_3}{}^+)^q - \left[(v_{a_1}{}^+)^q - (v_{a_3}{}^+)^q\right]\right|\end{array}\right\}$

$= \dfrac{1}{4}\left\{\begin{array}{l}(v_{a_1}{}^-)^q - (v_{a_3}{}^-)^q - \left[(u_{a_1}{}^-)^q - (u_{a_3}{}^-)^q\right] + \\ (v_{a_1}{}^+)^q - (v_{a_3}{}^+)^q - \left[(u_{a_1}{}^+)^q - (u_{a_3}{}^+)^q\right]\end{array}\right\}$

$$d(a_1,a_3) - d(a_1,a_2) = \frac{1}{4}\begin{bmatrix}(u_{a_3}^-)^q - (u_{a_2}^-)^q + (v_{a_2}^-)^q - (v_{a_3}^-)^q + \\ (u_{a_3}^+)^q - (u_{a_2}^+)^q + (v_{a_2}^+)^q - (v_{a_3}^+)^q\end{bmatrix} \geq 0$$

Therefore $d(a_1,a_2) \leq d(a_1,a_3)$, the same reason can be obtained $d(a_2,a_3) \leq d(a_1,a_3)$, so the property (4) can be proved.

In decision-making, reference point theory is often used to judge the views of decision-makers. Therefore, usually find the distance between the fuzzy number and the positive ideal point $<[1,1],[0,0]>$, and the negative ideal point $<[0,0],[1,1]>$. This article is inspired by Intuitionistic Fuzzy Positive Ideal Distance (PIS), Negative Ideal Distance (NIS), and Center-Selected Distance (CIS).

The distance between interval-valued generalized orthogonal fuzzy numbers and negative ideals and positive ideals is defined, and the compromise distance is defined.

**Definition 3.2:** The distance between any IVq-ROFNs $a = <[u^-, u^+], [v^-, v^+]>$ and the negative ideal $<[0,0],[1,1]>$ is called the interval-valued generalized orthogonal fuzzy negative ideal distance, as shown in formula (14).

$$NIS(a,0) = \frac{1}{4}(|(u^-)^q - (0)^q - ((v^-)^q - (1)^q)| + |(u^+)^q - (0)^q - ((v^+)^q - (1)^q)|) \quad (14)$$

**Definition 3.3:** The distance between any IVq-ROFNs $a = <[u^-, u^+], [v^-, v^+]>$ and the positive ideal $<[1,1],[0,0]>$ is called the interval-valued generalized orthogonal fuzzy positive ideal distance, as shown in formula (15).

$$PIS(a,1) = \frac{1}{4}(|(u^-)^q - (1)^q - ((v^-)^q - (0)^q)| + |(u^+)^q - (1)^q - ((v^+)^q - (0)^q)|) \quad (15)$$

**Definition 3.4:** The formula for the compromise distance of any two IVq-ROFNs is shown in (16).

$$CIS = \vartheta * PIS + (1 - \vartheta)NIS \quad (16)$$

In formula (16), $\vartheta$ satisfies: $0 \leq \vartheta \leq 1$。When $\vartheta = 0$，CIS was equal to *NIS*; when $\vartheta = 1$, CIS was equal to *PIS*; when $\vartheta > 0.5$ that the decision-maker's viewpoint was optimistic; when $\vartheta < 0.5$, that the decision-maker's viewpoint was conservative; and when $\vartheta = 0.5$, that the decision-maker's viewpoint was pertinent.

**Example 3.1:** Set $a_1 = <[0.85, 0.95], [0.1, 0.2]>$ as IVq-ROFN, and q=3, then according to formulas (14), (15), (16), Can get: $NIS(a,0) \approx 0.87$； $PIS(a,1) \approx 0.13$; when $\vartheta = 0.7$， $CIS \approx 0.35$, the result of the selection was close to *PIS*，indicating that the decision makers were optimistic. when $\vartheta = 0.2$，$CIS \approx 0.72$，the result of the selection was close to *NIS*, indicating that the decision-maker's viewpoint was conservative. when $\vartheta = 0.5$，$CIS \approx 0.5$, the result of the selection was between *PIS* and *NIS*, indicating that the decision-maker's point of view was pertinent.

## 3.2 Score function

In the MADM and MAGDM problems of intuitionistic fuzzy sets, the score function is used to measure the score value of each program, it is used for the ranking and selection of schemes, which has attracted the attention of researchers. For example: Literature [39], [30], [31] have given score functions good for sorting interval-valued intuitionistic fuzzy sets. However, they also have shortcomings as their definitions and shortcomings are given below.

**Definition 3.5** [29]: $a = <[u^-, u^+],[v^-, v^+]>$ is known to be an interval valued fuzzy number, the definition of the score function proposed by Cheng is shown in formula (17).

$$S_C(a) = \frac{1}{2}(u^- - v^- + u^+ - v^+) + 1 \qquad (17)$$

**Example 3.2:** Suppose $a_1 = <[0.15,0.2],[0.15,0.2]>$, $a_2 = <[0.3,0.4],[0.3,0.4]>$ are two IVq-IFNs, according to formula (17), we can know: $S_C(a_1) = 1, S_C(a_2) = 1$, then the score function of $S_C$ cannot compare these two IVq-IFNs.

**Definition 3.6** [30]: $a = <[u^-, u^+],[v^-, v^+]>$ is known to be an interval valued fuzzy number, the definition of the score function proposed by Bai is shown in formula (18).

$$S_B(a) = \frac{1}{2}[u^- + u^+ + u^-(1 - u^- - v^-) + u^+(1 - u^+ - v^+)] \qquad (18)$$

**Example 3.3:** Set $a_1 = <[0.5,0.5],[0.3,0.3]>$, $a_2 = <[0.4,0.4],[0.1,0.1]>$ as two IVq-IFNs, according to formula (18), we can get: $S_B(a_1) = 0.6, S_B(a_2) = 0.6$, then the score function of $S_B$ cannot compare these two IVq-IFNs.

**Definition 3.7** [31]: It is known that $a = <[u^-, u^+],[v^-, v^+]>$ is an interval valued fuzzy number, the definition of the score function of $S_{GM}(a)$ proposed by Gong and Ma, as shown in formula (19).

$$S_{GM}(a) = \frac{1}{2}(v^+ + v^- - u^+ - u^-) + \frac{u^+ + u^- + 2[(u^+)(u^-) - (v^+)(v^-)]}{v^+ + v^- + u^+ + u^-} \qquad (19)$$

**Example 3.4:** Suppose $a_1 = <[0.4,0.6],[0.4,0.6]>$, $a_2 = <[0.2,0.5],[0.2,0.5]>$ are two IVq-IFNs, according to formula (19), we can get: $S_{GM}(a_1) = 0.5, S_{GM}(a_2) = 0.5$, it can be seen that the score function of $S_{GM}$ cannot compare these two IVq-IFNs.

In order to overcome the shortcomings of the existing interval-valued fuzzy set scoring function, this paper draws inpiration from the interval value function proposed by Chen et al[28],

This paper proposes an interval-valued generalized orthogonal scoring function based on the principle of center selection. As shown in definition 3.8.

**Definition 3.8:** $a = <[u^-, u^+],[v^-, v^+]>$ is known to be an interval valued fuzzy number, The score function of the interval fuzzy number is shown in formula (20)( $q > 1, \alpha + \beta = 1, \alpha > 0, \beta > 0$）.

$$S(a) = \frac{1}{2}[\alpha(\sqrt[q]{u^-} + \sqrt[q]{1 - v^-}) + \beta(\sqrt[q]{u^+} + \sqrt[q]{1 - v^+})] \qquad (20)$$

For the sum of any two interval-valued fuzzy numbers such $a_1 = <[u_{a_1}^-, u_{a_1}^+],[v_{a_1}^-, v_{a_1}^+]>$ and $a_2 = <[u_{a_2}^-, u_{a_2}^+],[v_{a_2}^-, v_{a_2}^+]>$, formula (20) satisfies the following properties:

(1) If $a_1 > a_2$, then $S(a_1) > S(a_2)$; if $a_1 < a_2$, then $S(a_1) < S(a_2)$; if $a_1 = a_2$, then $S(a_1) = S(a_2)$

(2) If $0 \leq u_{a_1}^- \leq u_{a_1}^+ \leq 1; 0 \leq v_{a_1}^- \leq v_{a_1}^+ \leq 1; (u_{a_1}^+)^q + (v_{a_1}^+)^q \leq 1$, then $S(a_1) \in [0,1]$

(3) If $a_2 = <[0,0],[1,1]>$, then $S(a_2) = 0$

(4) If $a_1 = <[1,1],[0,0]>$, then $S(a_1) = 1$

(5) $S(a)$ aiming at $\alpha$ and $\beta$ having the same monotonicity

The properties (2), (3), and (4) of the score function are easy to prove. This article only gives the proof process of properties (1) and (5).

**Proof: Property (1):**

$$S(a_1) = \frac{1}{2}\left[\alpha\left(\sqrt[q]{u_{a_1}^-} + \sqrt[q]{1-v_{a_1}^-}\right) + \beta\left(\sqrt[q]{u_{a_1}^+} + \sqrt[q]{1-v_{a_1}^+}\right)\right]$$

$$S(a_2) = \frac{1}{2}\left[\alpha\left(\sqrt[q]{u_{a_2}^-} + \sqrt[q]{1-v_{a_2}^-}\right) + \beta\left(\sqrt[q]{u_{a_2}^+} + \sqrt[q]{1-v_{a_2}^+}\right)\right]$$

$$S(a_1)-S(a_2) = \frac{1}{2}\left[\begin{array}{l}\alpha\left(\sqrt[q]{u_{a_1}^-} - \sqrt[q]{u_{a_2}^-}\right) + \alpha\left(\sqrt[q]{1-v_{a_1}^-} - \sqrt[q]{1-v_{a_2}^-}\right) + \\ \beta\left(\sqrt[q]{u_{a_1}^+} - \sqrt[q]{u_{a_2}^+}\right) + \beta\left(\sqrt[q]{1-v_{a_1}^+} - \sqrt[q]{1-v_{a_2}^+}\right)\end{array}\right]$$

When $a_1 > a_2$, there is $u_{a_1}^- > u_{a_2}^-, u_{a_1}^+ > u_{a_2}^+, v_{a_1}^- < v_{a_2}^-, v_{a_1}^+ < v_{a_2}^+$, then there is $S(a_1)-S(a_2) > 0$, that is $S(a_1) > S(a_2)$;

When $a_1 < a_2$, there is $u_{a_1}^- < u_{a_2}^-, u_{a_1}^+ < u_{a_2}^+, v_{a_1}^- > v_{a_2}^-, v_{a_1}^+ > v_{a_2}^+$, then there is $S(a_1)-S(a_2) < 0$, that is $S(a_1) < S(a_2)$;

When $a_1 = a_2$, there is $u_{a_1}^- = u_{a_2}^-, u_{a_1}^+ = u_{a_2}^+, v_{a_1}^- = v_{a_2}^-, v_{a_1}^+ = v_{a_2}^+$, then there is $S(a_1)-S(a_2) = 0$, that is $S(a_1) = S(a_2)$;

The nature (1) is proved.

**Property (5):** derivation the parameter $\alpha$, get $S'_\alpha(a) = \frac{1}{2}\left(\sqrt[q]{(u^-)} + \sqrt[q]{(1-v^-)}\right)$, derivation the parameter $\beta$, get $S'_\beta(a) = \frac{1}{2}\left(\sqrt[q]{(u^+)} + \sqrt[q]{(1-v^+)}\right)$. It can be found, $S'_\alpha(a) > 0$, $S'_\beta(a) > 0$, $S(a)$ monotonically increase the parameters $\alpha$ and $\beta$ respectively. Therefore, the scoring function $S(a)$ has monotonicity for the parameters $\alpha$ and $\beta$ respectively, and the monotonicity is the same, and the property (5) is proved.

**Example 3.5:** For the three sets of data in Example 3.2, 3.3 and 3.4, according to the scoring function $S$ proposed in this article, we can get: $(q = 3, \alpha = 0.5, \beta = 0.5)$, if $a_1 =< [0.15,0.2],[0.15,0.2] >, a_2 =< [0.3,0.4],[0.3,0.4] >$, then $S(a_1) = 0.74793$, $S(a_2) = 0.78439$, the score values of the two numbers are different and can be compared; if $a_1 =< [0.5,0.5],[0.3,0.3] >$, $a_2 =< [0.4,0.4],[0.1,0.1] >$, then $S(a_1) = 0.84080$, $S(a_2) = 0.85115$, the score values of the two numbers are different, you can compare them; if $a_1 =< [0.4,0.6],[0.4,0.6] >$, $a_2 =< [0.2,0.5],[0.2,0.5] >$, then $S(a_1) = 0.79012$, $S(a_2) = 0.77513$, the score values of the two numbers are different, you can compare them.

**Definition 3.9:** let $a_1 =< [u_{a_1}^-, u_{a_1}^+],[v_{a_1}^-, v_{a_1}^+] >$ and $a_2 =< [u_{a_2}^-, u_{a_2}^+],[v_{a_2}^-, v_{a_2}^+] >$ are two IVq-IFNs, and the comparison of their sizes is as follows:

（1）　If $S(a_1) > S(a_2)$, then $a_1 > a_2$;

（2）　If $S(a_1) < S(a_2)$, then $a_1 < a_2$;

According to Example 3.5, it can be found that the score function proposed in this paper can be compared with the IVq-IFNs in Examples 3.2, 3.3, and 3.4. To further verify the scoring function proposed in this article, the interval value score function proposed in this paper is the same as the score function $S_C$ proposed by Cheng [29], the score function $S_B$ proposed by Bai [30], and the score function $S_{GM}$ proposed by Gong and Ma [31]. They were cited as examples with some randomly chosen interval-valued fuzzy numbers, and the comparison is shown in Table 1.

Table 1 Comparison results of each scoring function

| Interval fuzzy number($q > 1$) | Scoring function $S_C$ | Scoring function $S_{GM}$ | Scoring function $S_B$ | Proposed scoring function $S$ |
|---|---|---|---|---|
| $h_1$=([0.2, 0.6],[0.2, 0.4])<br>$h_2$=([0.3, 0.5],[0.1, 0.5]) | $h_1$=$h_2$ | $h_1$<$h_2$ | $h_1$<$h_2$ | $h_1$<$h_2$ |
| $h_3$=([0.3, 0.5],[0.2, 0.4])<br>$h_4$=([0.35,0.45],[0.25,0.35]) | $h_3$=$h_4$ | $h_3$=$h_4$ | $h_3$<$h_4$ | $h_3$<$h_4$ |
| $h_5$=([0.3, 0.6],[0.2, 0.3])<br>$h_6$=([0.4, 0.5],[0.1, 0.4]) | $h_5$=$h_6$ | $h_5$<$h_6$ | $h_5$<$h_6$ | $h_5$<$h_6$ |
| $h_7$=([0.1, 0.4],[0.2, 0.5])<br>$h_8$=([0.2, 0.3],[0.3, 0.4]) | $h_7$=$h_8$ | $h_7$=$h_8$ | $h_7$<$h_8$ | $h_7$<$h_8$ |
| $h_9$=([0.0, 0.0],[0.1, 0.1])<br>$h_{10}$=([0.0, 0.0],[0.9, 0.9]) | $h_9$>$h_{10}$ | $h_9$=$h_{10}$ | $h_9$=$h_{10}$ | $h_9$>$h_{10}$ |
| $h_{11}$=([0.6, 0.6],[0.4, 0.4])<br>$h_{12}$=([0.4, 0.4],[0.1, 0.1]) | $h_{11}$<$h_{12}$ | $h_{11}$<$h_{12}$ | $h_{11}$=$h_{12}$ | $h_{11}$<$h_{12}$ |

According to the comparison results in Table 1, it can be found that the score function proposed in this paper can not only make up for the shortcomings of other score functions, but also can accurately judge the size of the two interval value fuzzy numbers.

## 4. IVq-ROFYWA operator and IVq-ROFYWG operator

This part will define the generalized orthogonal Yager addition, multiplication, multiplication, and exponentiation algorithms and their properties in Section 4.1 according to the operating rules of the Yager operator in Section 2.2. In Section 4.2, define the Yager weighted average operator (IVq- ROFYWA), used for the weighted aggregation of multiple IVq-ROFNs, defines the Yager weighted geometric mean operator (IVq-ROFYWG) in Section 4.3, and integrates IVq-ROFYWA into the CRITIC and WASPAS methods.

### 4.1 IVq-ROFY operator

According to formulas (11) and (12) the basic algorithm of Yager operator, and the t-norms and t-conorms algorithm of fuzzy sets based on the Yager operator in Liu et al[14], this section begins to define the interval value Generalized orthogonal fuzzy Yager add, multiply, and power operation rules and give the related properties.

**Definition 4.1:** Let $a = [u^-, u^+], [v^-, v^+]$, $a_1 = [u_1^-, u_1^+], [v_1^-, v_1^+]$, $a_2 = [u_2^-, u_2^+], [v_2^-, v_2^+]$ be three interval-valued fuzzy numbers, and $\delta > 0$，interval-valued generalized orthogonal fuzzy Yager operator algorithm is shown in formulas (21), (22), (23), (24).

$$a_1 \oplus_Y a_2 = \left( \begin{array}{c} \left[ \sqrt[q]{min\left\{1,\left[(u_1{}^-)^{qp} + (u_2{}^-)^{qp}\right]^{\frac{1}{p}}\right\}}, \sqrt[q]{min\left\{1,\left[(u_1{}^+)^{qp} + (u_2{}^+)^{qp}\right]^{\frac{1}{p}}\right\}} \right], \\ \left[ \sqrt[q]{1 - min\left\{1,\left[\left(1-(v_1{}^-)^q\right)^p + \left(1-(v_2{}^-)^q\right)^p\right]^{\frac{1}{p}}\right\}}, \\ \sqrt[q]{1 - min\left\{1,\left[\left(1-(v_1{}^+)^q\right)^p + \left(1-(v_2{}^+)^q\right)^p\right]^{\frac{1}{p}}\right\}} \right] \end{array} \right) \quad (21)$$

$$a_1 \otimes_Y a_2 = \left( \begin{array}{c} \left[ \sqrt[q]{1 - min\left\{1,\left[\left(1-(u_1{}^-)^q\right)^p + \left(1-(u_2{}^-)^q\right)^p\right]^{\frac{1}{p}}\right\}}, \\ \sqrt[q]{1 - min\left\{1,\left[\left(1-(u_1{}^+)^q\right)^p + \left(1-(u_2{}^+)^q\right)^p\right]^{\frac{1}{p}}\right\}} \right], \\ \left[ \sqrt[q]{min\left\{1,\left[(v_1{}^-)^{qp} + (v_2{}^-)^{qp}\right]^{\frac{1}{p}}\right\}}, \sqrt[q]{min\left\{1,\left[(v_1{}^+)^{qp} + (v_2{}^+)^{qp}\right]^{\frac{1}{p}}\right\}} \right] \end{array} \right) \quad (22)$$

$$\delta a = \left( \begin{array}{c} \left[ \sqrt[q]{min\left\{1,\left[\delta(u^-)^{qp}\right]^{\frac{1}{p}}\right\}}, \sqrt[q]{min\left\{1,\left[\delta(u^+)^{qp}\right]^{\frac{1}{p}}\right\}} \right], \\ \left[ \sqrt[q]{1 - min\left\{1,\left[\delta\left(1-(v^-)^q\right)^p\right]^{\frac{1}{p}}\right\}}, \sqrt[q]{1 - min\left\{1,\left[\delta\left(1-(v^+)^q\right)^p\right]^{\frac{1}{p}}\right\}} \right] \end{array} \right) \quad (23)$$

$$a^\delta = \left( \begin{array}{c} \left[ \sqrt[q]{1 - min\left\{1,\left[\delta\left(1-(u^-)^q\right)^p\right]^{\frac{1}{p}}\right\}}, \sqrt[q]{1 - min\left\{1,\left[\delta\left(1-(u^+)^q\right)^p\right]^{\frac{1}{p}}\right\}} \right], \\ \left[ \sqrt[q]{min\left\{1,\left[\delta(v^-)^{qp}\right]^{\frac{1}{p}}\right\}}, \sqrt[q]{min\left\{1,\left[\delta(v^+)^{qp}\right]^{\frac{1}{p}}\right\}} \right] \end{array} \right) \quad (24)$$

It is easy to verify that the interval-valued generalized fuzzy Yager addition, multiplication, number multiplication and power proposed in this paper satisfy the basic definition of Yager operator.

**Theorem 3:** Let $a = [u^-, u^+], [v^-, v^+]$, $a_1 = [u_1^-, u_1^+], [v_1^-, v_1^+]$, $a_2 = [u_2^-, u_2^+], [v_2^-, v_2^+]$ be three interval-valued generalized orthogonal fuzzy numbers, and $\delta, \delta_1, \delta_2 > 0$, satisfies the following six properties, as shown in formulas (25) to (30).

(1) The commutative law of addition: $a_1 \oplus_Y a_2 = a_2 \oplus_Y a_1$ (25)

(2) The commutative law of multiplication: $a_1 \otimes_Y a_2 = a_2 \otimes_Y a_1$ (26)

(3) Multiplication distribution law: $\delta(a_1 \oplus_Y a_2) = \delta a_1 \oplus_Y \delta a_2$ (27)

(4) The distribution law of power: $(a_1 \otimes_Y a_2)^\delta = (a_2)^\delta \otimes_Y (a_1)^\delta (\delta > 0)$ (28)

(5) The law of additive distribution: $(\delta_1 + \delta_2)a = \delta_1 a \oplus_Y \delta_2 a$ (29)

(6) The associative law of power: $a^{\delta_1} \otimes_Y a^{\delta_2} = a^{(\delta_1 + \delta_2)} (\delta_1, \delta_2 > 0)$ (30)

The proof process of Theorem 3 can be proved by using the interval-valued generalized orthogonal fuzzy Yager algorithm above. This paper gives the proof of the multiplicative distribution law formula (27). Other properties are easy to derive. This paper does not give a detailed derivation process.

**Proof:** Formula (27):

According to formulas (21) and (23), the following formulas can be obtained:

$$\delta(a_1 \oplus_Y a_2) = \delta \begin{pmatrix} \left[ \sqrt[q]{min\left\{1, \left[(u_1^-)^{qp} + (u_2^-)^{qp}\right]^{\frac{1}{p}}\right\}}, \sqrt[q]{min\left\{1, \left[(u_1^+)^{qp} + (u_2^+)^{qp}\right]^{\frac{1}{p}}\right\}} \right], \\ \left[ \sqrt[q]{1 - min\left\{1, \left[\left(1-(v_1^-)^q\right)^p + \left(1-(v_2^-)^q\right)^p\right]^{\frac{1}{p}}\right\}}, \\ \sqrt[q]{1 - min\left\{1, \left[\left(1-(v_1^+)^q\right)^p + \left(1-(v_2^+)^q\right)^p\right]^{\frac{1}{p}}\right\}} \right] \end{pmatrix}$$

$$= \begin{pmatrix} \left[ \sqrt[q]{min\left\{1, \left[\delta(u_1^-)^{qp} + \delta(u_2^-)^{qp}\right]^{\frac{1}{p}}\right\}}, \sqrt[q]{min\left\{1, \left[\delta(u_1^+)^{qp} + \delta(u_2^+)^{qp}\right]^{\frac{1}{p}}\right\}} \right], \\ \left[ \sqrt[q]{1 - min\left\{1, \left[\delta\left(1-(v_1^-)^q\right)^p + \delta\left(1-(v_2^-)^q\right)^p\right]^{\frac{1}{p}}\right\}}, \\ \sqrt[q]{1 - min\left\{1, \left[\delta\left(1-(v_1^+)^q\right)^p + \delta\left(1-(v_2^+)^q\right)^p\right]^{\frac{1}{p}}\right\}} \right] \end{pmatrix}$$

$$\delta a_1 \oplus_Y \delta a_2 = \begin{pmatrix} \left[ \sqrt[q]{min\left\{1, \left[\delta(u_1^-)^{qp}\right]^{\frac{1}{p}}\right\}}, \sqrt[q]{min\left\{1, \left[\delta(u_1^+)^{qp}\right]^{\frac{1}{p}}\right\}} \right], \\ \left[ \sqrt[q]{1 - min\left\{1, \left[\delta\left(1-(v_1^-)^q\right)^p\right]^{\frac{1}{p}}\right\}}, \sqrt[q]{1 - min\left\{1, \left[\delta\left(1-(v_1^+)^q\right)^p\right]^{\frac{1}{p}}\right\}} \right] \end{pmatrix} \oplus_Y$$

$$\begin{pmatrix} \left[ \sqrt[q]{min\left\{1, \left[\delta(u_2^-)^{qp}\right]^{\frac{1}{p}}\right\}}, \sqrt[q]{min\left\{1, \left[\delta(u_2^+)^{qp}\right]^{\frac{1}{p}}\right\}} \right], \\ \left[ \sqrt[q]{1 - min\left\{1, \left[\delta\left(1-(v_2^-)^q\right)^p\right]^{\frac{1}{p}}\right\}}, \sqrt[q]{1 - min\left\{1, \left[\delta\left(1-(v_2^+)^q\right)^p\right]^{\frac{1}{p}}\right\}} \right] \end{pmatrix}$$

$$= \left( \begin{array}{c} \left[ \sqrt[q]{min\left\{1, [\delta(u_1{}^-)^{qp} + \delta(u_2{}^-)^{qp}]^{\frac{1}{p}}\right\}}, \sqrt[q]{min\left\{1, [\delta(u_1{}^+)^{qp} + \delta(u_2{}^+)^{qp}]^{\frac{1}{p}}\right\}} \right], \\ \left[ \sqrt[q]{1 - min\left\{1, [\delta(1-(v_1{}^-)^q)^p + \delta(1-(v_2{}^-)^q)^p]^{\frac{1}{p}}\right\}}, \\ \sqrt[q]{1 - min\left\{1, [\delta(1-(v_1{}^+)^q)^p + \delta(1-(v_2{}^+)^q)^p]^{\frac{1}{p}}\right\}} \right] \end{array} \right)$$

$$= \delta(a_1 \oplus_Y a_2)$$

The formula （27） $\delta(a_1 \oplus_Y a_2) = \delta a_2 \oplus_Y \delta a_1$ is proved.

The operating rules and laws of Yager operator on interval-valued generalized orthogonal fuzzy sets lay the foundation for the Yager weighted average operator and its related theorems proposed in Section 4.2 of this paper.

## 4.2  IVq-ROFYWA operator

In order to better integrate the CRITIC and WASPAS methods, according to the generalized value orthogonal fuzzy sets of the Yager operator interval defined in Section 4.1, the addition, multiplication, number multiplication, and power operation rules, and the generalized orthogonality proposed by Liu et al. [14] The Yager weighted average formula of fuzzy numbers. This section begins to define the interval-valued generalized orthogonal Yager weighted average operator to aggregate and give related properties, which can be used to aggregate multiple interval-valued generalized fuzzy numbers.

**Definition 4.2:** Let $a_i = <[u_i{}^-, u_i{}^+], [v_i{}^-, v_i{}^+]>$ （i=1,2,3,…n） be a set of IVq-ROFNs, $\omega = (\omega_1, \omega_2, …, \omega_n)^T$ is a weight vector, and satisfy: $\sum_{i=1}^{n} \omega_i = 1, \omega_i \in [0,1]$, $(i = 1,2,…,n)$。The interval value generalized orthogonal Yager weighted average operator (IVq-ROFYWA) is defined as shown in formula (31).

$$\text{IVq - ROFYWA}(a_1, a_2, …, a_n) = \bigoplus_{i=1}^{n} \omega_i a_i \qquad (31)$$

**Theorem 4:** Let $a_i = <[u_i{}^-, u_i{}^+], [v_i{}^-, v_i{}^+]>$ （i=1,2,3,…n） be a set of IVq-ROFNs, and the aggregation value of IVq-ROFYWA is also a generalized orthogonal fuzzy number. The aggregation result is shown in formula (32).

$$\begin{aligned} &IVq - R0FYWA(a_1, a_2, …, a_n) \\ &= \left( \begin{array}{c} \left[ \sqrt[q]{min\left\{1, [\sum_{i=1}^{n} \omega_i (u_i{}^-)^{qp}]^{\frac{1}{p}}\right\}}, \sqrt[q]{min\left\{1, [\sum_{i=1}^{n} \omega_i (u_i{}^+)^{qp}]^{\frac{1}{p}}\right\}} \right], \\ \left[ \sqrt[q]{1 - min\left\{1, [\sum_{i=1}^{n} \omega_i (1-(v_i{}^-)^q)^p]^{\frac{1}{p}}\right\}}, \sqrt[q]{1 - min\left\{1, [\sum_{i=1}^{n} \omega_i (1-(v_i{}^+)^q)^p]^{\frac{1}{p}}\right\}} \right] \end{array} \right) \end{aligned} \qquad (32)$$

**Proof:** When n=2,

$$\omega_i a_i = \left( \begin{array}{c} \left[ \sqrt[q]{min\left\{1,\left[\omega_i(u_i^-)^{qp}\right]^{\frac{1}{p}}\right\}}, \sqrt[q]{min\left\{1,\left[\omega_i(u_i^+)^{qp}\right]^{\frac{1}{p}}\right\}} \right], \\ \left[ \sqrt[q]{1 - min\left\{1,\left[\omega_i(1-(v_i^-)^q)^p\right]^{\frac{1}{p}}\right\}}, \sqrt[q]{1 - min\left\{1,\left[\omega_i(1-(v_i^+)^q)^p\right]^{\frac{1}{p}}\right\}} \right] \end{array} \right)$$

$IVq - R0FYWA(a_1, a_2) = \omega_1 a_1 \oplus \omega_2 a_2$

$$= \left( \begin{array}{c} \left[ \sqrt[q]{min\left\{1,\left[\omega_1(u_1^-)^{qp} + \omega_2(u_2^-)^{qp}\right]^{\frac{1}{p}}\right\}}, \sqrt[q]{min\left\{1,\left[\omega_1(u_1^+)^{qp} + \omega_2(u_2^+)^{qp}\right]^{\frac{1}{p}}\right\}} \right], \\ \left[ \sqrt[q]{1 - min\left\{1,\left[\omega_1(1-(v_1^-)^q)^p + \omega_2(1-(v_2^-)^q)^p\right]^{\frac{1}{p}}\right\}}, \\ \sqrt[q]{1 - min\left\{1,\left[\omega_1(1-(v_1^+)^q)^p + \omega_2(1-(v_2^+)^q)^p\right]^{\frac{1}{p}}\right\}} \right] \end{array} \right)$$

$$= \left( \begin{array}{c} \left[ \sqrt[q]{min\left\{1,\left[\sum_{i=1}^{2}\omega_i(u_i^-)^{qp}\right]^{\frac{1}{p}}\right\}}, \sqrt[q]{min\left\{1,\left[\sum_{i=1}^{2}\omega_i(u_i^+)^{qp}\right]^{\frac{1}{p}}\right\}} \right], \\ \left[ \sqrt[q]{1 - min\left\{1,\left[\sum_{i=1}^{2}\omega_i(1-(v_i^-)^q)^p\right]^{\frac{1}{p}}\right\}}, \sqrt[q]{1 - min\left\{1,\left[\sum_{i=1}^{2}\omega_i(1-(v_i^+)^q)^p\right]^{\frac{1}{p}}\right\}} \right] \end{array} \right) \quad (33)$$

From equation (33), we can see that $IVq - ROFYWA(a_1, a_2) = \omega_1 a_1 \oplus_Y \omega_2 a_2$, when n=2, the equation (32) is satisfied.

Assuming that when n=k, the formula (32) is satisfied, then:

$IVq - R0FYWA(a_1, a_2, ..., a_k) = \oplus_{i=1}^{k} \omega_i a_i$

$$= \left( \begin{array}{c} \left[ \sqrt[q]{min\left\{1,\left[\sum_{i=1}^{k}\omega_i(u_i^-)^{qp}\right]^{\frac{1}{p}}\right\}}, \sqrt[q]{min\left\{1,\left[\sum_{i=1}^{k}\omega_i(u_i^+)^{qp}\right]^{\frac{1}{p}}\right\}} \right], \\ \left[ \sqrt[q]{1 - min\left\{1,\left[\sum_{i=1}^{k}\omega_i(1-(v_i^-)^q)^p\right]^{\frac{1}{p}}\right\}}, \sqrt[q]{1 - min\left\{1,\left[\sum_{i=1}^{k}\omega_i(1-(v_i^+)^q)^p\right]^{\frac{1}{p}}\right\}} \right] \end{array} \right)$$

Then when n=k+1, there are:

$IVq - R0FYWA(a_1, a_2, ..., a_{k+1}) = \left(\oplus_{i=1}^{k} \omega_i a_i\right) \oplus \omega_{k+1} a_{k+1}$

$$= \left( \begin{array}{c} \left[ \sqrt[q]{min\left\{1,\left[\sum_{i=1}^{k}\omega_i(u_i^{-})^{qp}\right]^{\frac{1}{p}}\right\}}, \sqrt[q]{min\left\{1,\left[\sum_{i=1}^{k}\omega_i(u_i^{+})^{qp}\right]^{\frac{1}{p}}\right\}} \right], \\ \left[ \sqrt[q]{1-min\left\{1,\left[\sum_{i=1}^{k}\omega_i\left(1-(v_i^{-})^q\right)^p\right]^{\frac{1}{p}}\right\}}, \sqrt[q]{1-min\left\{1,\left[\sum_{i=1}^{k}\omega_i\left(1-(v_i^{+})^q\right)^p\right]^{\frac{1}{p}}\right\}} \right] \end{array} \right) \oplus_Y$$

$$\left( \begin{array}{c} \left[ \sqrt[q]{min\left\{1,\left[\omega_{k+1}(u_{k+1}^{-})^{qp}\right]^{\frac{1}{p}}\right\}}, \sqrt[q]{min\left\{1,\left[\omega_{k+1}(u_{k+1}^{+})^{qp}\right]^{\frac{1}{p}}\right\}} \right], \\ \left[ \sqrt[q]{1-min\left\{1,\left[\omega_{k+1}\left(1-(v_{k+1}^{-})^q\right)^p\right]^{\frac{1}{p}}\right\}}, \sqrt[q]{1-min\left\{1,\left[\omega_{k+1}\left(1-(v_{k+1}^{+})^q\right)^p\right]^{\frac{1}{p}}\right\}} \right] \end{array} \right)$$

$$= \left( \begin{array}{c} \left[ \sqrt[q]{min\left\{1,\left[\sum_{i=1}^{k+1}\omega_i(u_i^{-})^{qp}\right]^{\frac{1}{p}}\right\}}, \sqrt[q]{min\left\{1,\left[\sum_{i=1}^{k+1}\omega_i(u_i^{+})^{qp}\right]^{\frac{1}{p}}\right\}} \right], \\ \left[ \sqrt[q]{1-min\left\{1,\left[\sum_{i=1}^{k+1}\omega_i\left(1-(v_i^{-})^q\right)^p\right]^{\frac{1}{p}}\right\}}, \sqrt[q]{1-min\left\{1,\left[\sum_{i=1}^{k+1}\omega_i\left(1-(v_i^{+})^q\right)^p\right]^{\frac{1}{p}}\right\}} \right] \end{array} \right)$$

（34）

It can be seen from equation (34) that when n=k+1, equation (32) is also satisfied. Therefore, Theorem 4 is proved.

The interval-valued generalized orthogonal Yager weighted average IVq-ROFYWA operator also satisfies idempotence, boundedness, invariant interchangeability, and monotonicity. Its properties are given below.

**Theorem 5 (idempotence):** Suppose $a_i = <[u_i^{-},u_i^{+}],[v_i^{-},v_i^{+}]>, a = <[u^{-},u^{+}],[v^{-},v^{+}]>$ （i=1,2,….,n）is a set of IVq-ROFNs, and $a_i = a$, then there is formula (35):

$$IVq - ROFYWA(a_1,a_2,a_3,...a_n) = a \qquad （35）$$

**Theorem 6 (boundedness):** Let $a_i = <[u_i^{-},u_i^{+}],[v_i^{-},v_i^{+}]>$ （i=1,2,….,n）be a set of IVq-ROFNs, then formula (36):

$$a_{min} \leq IVq - ROFYWA \leq a_{max} \qquad （36）$$

$$a_{max} = \max(a_i) \qquad （37）$$

$$a_{min} = \min(a_i) \qquad （38）$$

**Theorem 7 (monotonicity):** Suppose: $a_i = <[u_i^{-},u_i^{+}],[v_i^{-},v_i^{+}]>, a_i^* = <[u^*{}_i^{-},u^*{}_i^{+}],[v^*{}_i^{-},v^*{}_i^{+}]>$ （i=1,2,….,n）are two sets of IVq-ROFNs, and satisfy $a_i \leq a_i^*$, then there is formula (39):

$$IVq - RVOYWA(a_1,a_2,a_3,...a_n) \leq IVq - RVOYWA(a_1^*,a_2^*,a_3^*,...a_n^*) \qquad （39）$$

**Theorem 8 (Permutation Invariance):** Suppose $(\alpha_1',\alpha_2',\cdots,\alpha_n')$ is a set of IVq-ROFNs, and is any permutation of $(\alpha_1,\alpha_2,\cdots,\alpha_n)$, then:

$$IVq - RVOYWA(\alpha_1,\alpha_2,\cdots,\alpha_n) = IVq - RVOYWA(\alpha_1',\alpha_2',\cdots,\alpha_n') \qquad （40）$$

Using formulas (21), (22), (23), (24), it is easy to prove idempotence, boundedness,

monotonicity, and permutation invariance. This article does not give a detailed proof process.

This section proposes the Yager weighted average operator (IVq-ROFYWA) in the interval-value generalized orthogonal environment, which can realize the aggregation of interval-valued data of multiple decision makers, and will be integrated with the decision-making method and group decision-making method proposed in this paper.

## 4.3 IVq-ROFYGM operator

According to the algorithm of addition, multiplication, number multiplication, and power operation of the generalized value orthogonal fuzzy set of the Yager operator interval defined in Section 4.1, and the Yager weighted geometric average operator of generalized orthogonal fuzzy numbers proposed by Liu et al. [14], This paper proposes interval-valued generalized orthogonal Yager weighted geometric mean operators and their properties.

**Definition 4.3:** Let $a_i = <[u_i^-,u_i^+],[v_i^-,v_i^+]>$ (i = 1,2,3,...n) be a set of IVq-ROFNs, $\omega = (\omega_1,\omega_2,...,\omega_n)^T$ be a weight vector, and satisfy: $\sum_{i=1}^{n}\omega_i = 1, \omega_i \in [0,1]$, $(i = 1,2,...,n)$。 The interval value generalized orthogonal Yager weighted average operator (IVq-ROFYWG) is defined as shown in formula (41).

$$IVq\text{-}ROFYWG(a_1,a_2,...,a_n) = \otimes_{i=1}^{n} a_i^{\omega_i} \quad (41)$$

**Theorem 9:** Let $a_i = <[u_i^-,u_i^+],[v_i^-,v_i^+]>$ （i=1,2,3,…n） be a set of IVq-ROFNs, and the aggregation value of IVq-ROFYWA is also a generalized orthogonal fuzzy number. The aggregation result is shown in formula (42).

$$IVq - ROFYWG(a_1,a_2,...,a_n)$$

$$= \left( \begin{array}{c} \left[\sqrt[q]{1 - min\left\{1,\left[\sum_{i=1}^{n}\omega_i\left(1-(u_i^-)^q\right)^p\right]^{\frac{1}{p}}\right\}}, \sqrt[q]{1 - min\left\{1,\left[\sum_{i=1}^{n}\omega_i\left(1-(u_i^+)^q\right)^p\right]^{\frac{1}{p}}\right\}}\right], \\ \left[\sqrt[q]{min\left\{1,\left[\sum_{i=1}^{n}\omega_i(v_i^-)^{qp}\right]^{\frac{1}{p}}\right\}}, \sqrt[q]{min\left\{1,\left[\sum_{i=1}^{n}\omega_i(v_i^+)^{qp}\right]^{\frac{1}{p}}\right\}}\right] \end{array} \right)$$

（42）

The proof process is similar to Theorem 4. This article does not give a detailed proof.

**Theorem 10:** The interval-valued generalized orthogonal Yager weighted geometric average IVq-ROFYWG operator also satisfies: (1) idempotence; (2) boundedness; (3) permutation invariance; (4) monotonicity.

(1) **Idempotence:** Let $a_i = <[u_i^-,u_i^+],[v_i^-,v_i^+]>, a = <[u^-,u^+],[v^-,v^+]>$ （i=1,2,….,n）be a set of IVq-ROFNs, satisfying $a_i = a$, then there is formula (43):

$$IVq - ROFYWG(a_1,a_2,a_3,...a_n) = a \quad (43)$$

(2) **Boundedness:** Let (i=1,2,...,n) be a set of IVq-ROFNs, then formula (44):

$$a_{min} \leq IVq - ROFYWG \leq a_{max} \quad (44)$$

$$a_{max} = \max(a_i) \quad (45)$$

$$a_{min} = \min(a_i) \quad (46)$$

(3) **Monotonicity:** Let: $a_i = <[u_i^-,u_i^+],[v_i^-,v_i^+]>, a_i^* = <[u^*_i{}^-,u^*_i{}^+],[v^*_i{}^-,v^*_i{}^+$

] > (i = 1,2,...,n) be two sets of IVq-ROFNs, and satisfy: $a_i \leq a_i^*$, then there is formula (47):

$$IVq - ROFYWG(a_1,a_2,a_3,...a_n) \leq IVq - ROFYWG(a_1^*,a_2^*,a_3^*,...a_n^*) \quad (47)$$

(4) **Permutation invariance:** Let $(\alpha'_1,\alpha'_2,\cdots,\alpha'_n)$ be a set of IVq-ROFNs and any permutation of $(\alpha_1,\alpha_2,\cdots,\alpha_n)$, then there is formula (48):

$$IVq - RVOYWG(\alpha_1,\alpha_2,\cdots,\alpha_n) = IVq - RVOYWG(\alpha'_1,\alpha'_2,\cdots,\alpha'_n) \quad (48)$$

Using formulas (21), (22), (23), (24), it is easy to prove the idempotence, boundedness, monotonicity, and permutation invariance of the IVq-ROFYWG operator. This article does not give a detailed proof process.

This section proposes the Yager weighted geometric average operator (IVq-ROFYWG) in the interval-value generalized orthogonal environment, which will be applied in the CRITIC-WASPAS fusion decision method described in this article.

## 5. CRTIC-WASPAS decision method

This part describes the operating environment of the decision-making method in 5.1, and then extends the CRITIC method and WASPAS method to interval-valued generalized orthogonal fuzzy in sections 5.2 and 5.3, and then performs interval-valued generalized orthogonal fuzzy in section 5.4 The fusion of CRITIC and WASPAS methods.

## 5.1 Decision environment description

The calculation of CRITIC method and WASPAS method is based on the decision matrix to calculate the attribute weights and determine the order of the schemes. It is known that $Y = \{y_1, y_2,...y_m\}$ is the scheme set, $C = \{c_1,c_2,...c_n\}$ is the attribute set. A decision matrix can be obtained by combining different schemes and attribute values. The attribute weight $\omega_j$ is unknown, but it satisfies: $\sum_{j=1}^{n}\omega_j = 1$. The decision matrix is expressed as: $A = [a_{ij}]_{m \times n}$ as shown in (49), where $a_{ij} = <[u_{a_{ij}}^-,u_{a_{ij}}^+],[v_{a_{ij}}^-,v_{a_{ij}}^+]>$, $a_{ij}$ is IVq-ROFNs, satisfying $(u_{a_{ij}}^+)^q + (v_{a_{ij}}^+)^q \leq 1$, $(u_{a_{ij}}^-)^q + (v_{a_{ij}}^-)^q \leq 1$, so that a suitable q value can be determined.

$$A = \begin{bmatrix} <[u_{a_{11}}^-,u_{a_{11}}^+],[v_{a_{11}}^-,v_{a_{11}}^+]> & \cdots & <[u_{a_{1n}}^-,u_{a_{1n}}^+],[v_{a_{1n}}^-,v_{a_{1n}}^+]> \\ \vdots & \ddots & \vdots \\ <[u_{a_{m1}}^-,u_{a_{m1}}^+],[v_{a_{m1}}^-,v_{a_{m1}}^+]> & \cdots & <[u_{a_{mn}}^-,u_{a_{mn}}^+],[v_{a_{mn}}^-,v_{a_{mn}}^+]> \end{bmatrix} \quad (49)$$

In the decision-making process, there are benefit type $\Omega\_1$ and cost type $\Omega\_2$ according to the characteristics of attributes. At this time, the matrix needs to be standardized. Many processing methods are used in this paper to deal with formula (50).

$$x_{ij} = \begin{cases} \frac{a_{ij} - min\,(a_{ij})}{max\,(a_{ij}) - min\,(a_{ij})}, & a_{ij} \in \Omega_1 \\ \frac{max\,(a_{ij}) - a_{ij}}{max\,(a_{ij}) - min\,(a_{ij})}, & a_{ij} \in \Omega_2 \end{cases} ;(i = 1,2,...,m; j = 1,2,...,n) \quad (50)$$

The CRITIC-WASPAS decision-making method given in this paper is used to solve the multi-attribute decision-making problem with unknown attribute weights. In this fusion method, the weight of each attribute is first determined by the interval value generalized orthogonal CRITIC method, and then the interval value generalized orthogonal WASPAS method is used to calculate the final interval value score of each program, and finally by comparing the interval of each program The value score determines the ranking of alternatives and the optimal solution.

## 5.2 Interval-valued generalized orthogonal fuzzy CRITIC method

The CRITIC method was proposed by Diakoulaki, Mavrotas, and Papayannakis in 1995[15]. When calculating the weights, the CRITIC method not only considers the impact of variation on the indicators, but also the impact of correlation on the indicators. This method can eliminate the influence of some indicators with strong correlation, reduce the overlap of information between indicators, and help obtain credible evaluation results. Due to the advantages of the CRITIC method, it has attracted the attention of researchers. Among them, Peng et al.[35] proposed a PFS-based TOPSIS-CRITIC decision method. The researchers extended it to intuitionistic fuzzy sets. Lai and Liao proposed a method based on DNMA and CRITIC language D value Multi-criteria decision-making methods are used for blockchain platform evaluation. Inspired by [15], this paper proposes the CRITIC method of interval-valued generalized orthogonal fuzzy numbers to calculate attribute weights. The CRITIC method includes 5 steps: (1) Standardize the decision matrix; (2) Calculate the correlation coefficient between each attribute; (3) Calculate the attribute standard deviation of the correlation coefficient between each attribute; (4) Calculate the index; (5) Get the weight of each attribute. The realization of the CRITIC method under the condition of interval value generalized orthogonal fuzzy requires calculation: the mean value, correlation coefficient, standard deviation, attribute index value of multiple IVq-ROFNs, where the mean value adopts the IVq-ROFYWA operator, and the other definitions will be given below.

Definition 5.1: Suppose a_i and b_i are two sets of IVq-ROFNs (i=1,2,...,n), then their correlation coefficient ρ is shown in formula (51).

$$\rho = \{ \oplus_{i=1}^{n}[(a_i \ominus \bar{a}) \otimes (b_i \ominus \bar{b})]\} \oslash \left\{[(\oplus_{i=1}^{n}(a_i \ominus \bar{a})^2) \otimes (\oplus_{i=1}^{n}(b_i \ominus \bar{b})^2)]^{\frac{1}{2}}\right\} \quad (51)$$

In formula (43), $\bar{a}$ and $\bar{b}$ respectively represent the mean value of two sets of interval-valued generalized orthogonal fuzzy numbers of $a_i$ and $b_i$.

Definition 5.2: Suppose a_i is a set of IVq-ROFNs (i=1,2,...,n), then the standard deviation σ is calculated as shown in formula (52).

$$\sigma = \left[\frac{1}{n} \oplus_{i=1}^{n} (a_i \ominus \bar{a})^2\right]^{\frac{1}{2}} \quad (52)$$

The specific calculation steps of the interval value generalized orthogonal fuzzy CRITIC method to determine the attribute weight according to the decision matrix A are as follows:

(1) The matrix standardization process uses formula (50).

(2) Solve the correlation coefficient. According to formula (53), the correlation coefficient ρ_jk between every two attributes can be determined, where ρ_jk is IVq-ROFNs.

$$\rho_{jk} = \frac{\sum_{i=1}^{m}(x_{ij} - \bar{x}_j)(x_{ik} - \bar{x}_k)}{\sqrt{\sum_{i=1}^{m}(x_{ij} - \bar{x}_j)^2 \sum_{i=1}^{m}(x_{ik} - \bar{x}_k)^2}} \quad (53)$$

In formula (54), $\bar{x}_j$ and $\bar{x}_k$ show the average value of the jth and kth attributes. Respectively calculated by formula (54), the weight in formula (55) $\omega_j = \frac{1}{n}(j = 1,2,...,n)$.

$$\bar{x}_j = IVq - ROFYWA(a_1, a_2, a_3, ... a_n); (j = 1,2,...,n) \quad (54)$$

(3) Solve for the standard deviation $\sigma_j$ (j=1,2,…,n). The standard deviation of each attribute

is estimated by formula (55), where $\sigma_j$ is IVq-ROFNs.

$$\sigma_j = \sqrt{\frac{1}{m}\Sigma_{i=1}^m \left(x_{ij} - \bar{x}_j\right)^2};(j = 1,2,...,n) \quad (55)$$

(4)Solve the index value $N_j$ (j=1,2,...,n) of each attribute. Formula (56) can determine the index of each attribute through the standard deviation of the attributes and the correlation coefficient between the attributes. At this time, $N_j$ is IVq-ROFNs.

$$N_j = \sigma_j \Sigma_{k=1}^n (1 - \rho_{jk});(j = 1,2,...,n) \quad (56)$$

(5)The index is normalized to obtain the interval value attribute weight. Use formula (57) to normalize the index of each attribute, where $w_j$ is IVq-ROFNs.

$$w_j = \frac{N_j}{\Sigma_{i=1}^n N_j};(j = 1,2,...,n) \quad (57)$$

CRITIC can handle the situation where the attribute weight is unknown, and then choosing the best from multiple schemes requires a better method. The best selection and ranking methods are given in the next section.

## 5.3 interval-valued generalized orthogonal fuzzy WASPAS method

The WASPAS**Error! Reference source not found.** method is based on the fusion of the weighted sum model (WSM) and the weighted product model (WPM). It overcomes the shortcomings of complex multiplication operations and increases the accuracy of the data, providing more accurate results. It is an effective MAGDM method. Researchers put forward new studies on the basis of WASPAS. For example, Katarzyna Rudnik et al. **Error! Reference source not found.** proposed the OFN-WASPAS method, and R. Davoudabadi et al.**Error! Reference source not found.** proposed the WASPAS-TOPSIS decision method on interval-valued intuitionistic fuzzy sets. Inspired by [16], this paper proposes the WASPAS method of interval-valued generalized orthogonal fuzzy numbers to determine the order of the schemes. The WASPAS method consists of 5 steps: (1) Standardized decision matrix; (2) The relative importance of the attributes of the weighted sum; (3) The relative importance of the attributes of the weighted product; (4) The joint generalized criterion, that is, the average of the weighted sum and the weighted product, (5) Sorting . The specific calculation steps of the interval-valued generalized orthogonal fuzzy WASPAS method to determine the attribute weights according to the decision matrix A are as follows:

（1）To standardize the decision matrix, use formula (50).

（2）The relative importance of the weighted sum $Q_i^{(1)}$. Formula (58) is used to determine the relative importance of the sum of weighted standardized data for each alternative, where $\omega_j$ （j=1,2,…,n）is the real weight of the attribute, at this time $Q_i^{(1)}$ are IVq-ROFNs.

$$Q_i^{(1)} = IVq - ROFYWA(a_1,a_2,a_3,...a_n);(i = 1,2,...,m) \quad (58)$$

（3）The relative importance of the weighted product $Q_i^{(1)}$. Formula (59) is used to determine the relative importance of the product of weighted standardized data for each alternative, where $\omega_j$ （j=1,2,…,n）is the weight of the attribute, at this time $Q_i^{(1)}$ are IVq-ROFNs.

$$Q_i^{(2)} = IVq - ROFYWG(a_1,a_2,a_3,...a_n);(i = 1,2,...,m) \quad (59)$$

（4）Find the weighted average $r_i$ of the relative importance of the weighted sum and the relative importance of the weighted product. Formula (60) is used to calculate the weighted average $r_i$（i=1,2,…,m） of the relative importance of the weighted sum and the relative importance of the weighted product, where $r_i$ is IVq-ROFNs。

$$r_i = \lambda Q_i^{(1)} + (1-\lambda)Q_i^{(2)}; (i=1,2,\ldots,m; 0 < \lambda < 1) \qquad (60)$$

In formula (60), when $\lambda = 1$, the formula will be converted to WSM model; when $\lambda = 0$, the formula will be converted to WSM model.

（5）Scheme sorting. Sort $r_i$ to get the order of the scheme.

The extended ASPAS method described in this paper is to solve the MADM problem of interval-valued generalized orthogonal fuzzy with known attribute weights. For the unknown weight problem, this paper will combine the WASPAS and CRITIC methods.

## 5.4 CRTIC-WASPAS MADM method

In order to better solve the problem of unknown attribute weights in the group decision , this section starts to study the advantages of the fusion of CRITIC and WASPAS methods, and proposes the interval value generalized orthogonal fuzzy CRITIC-WASPAS method. According to section 4.2, the CRITIC method is known Generalized to the interval-valued generalized orthogonal fuzzy centralized operation, the weight $w_j$ obtained is still the interval-valued generalized orthogonal fuzzy number. In order to better combine the CRITIC method and the WASPAS method, it is thought that the distance can reflect the two interval-valued generalized The difference of orthogonal fuzzy numbers and the theory of reference points, therefore, the ideal distance measurement method can be chosen to convert the weights of interval-valued generalized orthogonal fuzzy numbers into real number weights. According to the analysis of the three ideal distances in Section 3.1, the negative ideal distance is selected to realize the real number processing of the interval-valued generalized orthogonal fuzzy number weight $w_j$ of each attribute obtained by the CRITIC method. The specific calculation steps of the interval-valued generalized orthogonal fuzzy CRITIC-WASPAS method are as follows:

（1）Determine the value of q. $a_{ij}$ is an interval-valued generalized orthogonal fuzzy number, which satisfies $(u_{a_{ij}}^+)^q + (v_{a_{ij}}^+)^q \leq 1$, $(u_{a_{ij}}^-)^q + (v_{a_{ij}}^-)^q \leq 1$. When the amount of data is not large, the value of q can be determined by the observation method, and when the amount of data is large, it is necessary to find the appropriate value of q through the traversal test.

（2）Determine the interval value weight of each attribute. Input the decision matrix and the column of positive and negative attributes into the interval value generalized orthogonal fuzzy CRITIC method. After the CRITIC method, the interval value weight $w_j$（j=1,2,…,n） of each attribute is obtained.

（3）Determine the real weight of each attribute $w_j$（j=1,2,…,n） Normalize the negative ideal distance of $w_j$（j=1,2,…,n） (formula 14) $d_j$（j=1,2,…,n）, as shown in formula (61).

$$\omega_j = \frac{d_j}{\sum_{j=1}^{n} d_j} \qquad (61)$$

（4）Determine the final interval value score of the alternatives. Input the decision matrix, the positive and negative attribute columns and the weight of each attribute determined in step (3) $\omega = \{\omega_1, \omega_2, \ldots, \omega_n\}$ into the interval-valued generalized orthogonal fuzzy WASPAS method, which is calculated by the WASPAS method. The interval value score for each alternative.

（5）Sort the plans and choose the best plan. The size comparison between two interval-valued generalized orthogonal fuzzy numbers can be compared using the scoring function (Equation 20) defined in this article in Section 3.2.

The operation steps of the interval-valued generalized orthogonal fuzzy fusion CRITIC-WASPAS decision method are shown in Figure 1:

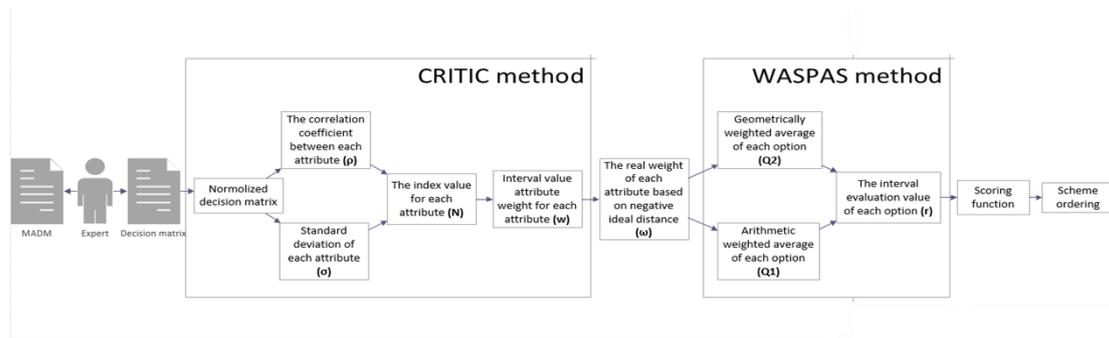

Figure 1 Flow chart of MADM method based on CRITIC-WASPAS

This part combines CRITIC and WASPAS methods on the basis of interval generalized orthogonality. By combining the advantages of the two, the decision-making problem of unknown attribute weight can be better solved. However, the fusion method can only complete the selection of the optimal scheme of one decision matrix. If you want to deal with the comprehensive optimal scheme selection of multiple decision matrices at the same time, you need to integrate the collective settlement operator of Part 4. The sixth part of this article will explain in detail how the entire group decision-making method completes the decision-making problem of multiple decision makers and multiple attributes without any weight.

# 6. Group decision-making method based on CRITIC-WASPAS

Based on the IVq-ROFYWA operator in part 4 and the interval-valued generalized orthogonal fuzzy CRITIC-WASPAS decision method in part 5, this part will elaborate on the group decision-making method proposed in this paper. Among them, the environment of the group decision-making method is given in Section 6.1, the method for solving the weight of decision-makers is given in Section 6.2, and the group decision-making algorithm and its flowchart are given in Section 6.3.

## 6.1 Group decision environment

In production and life, it is often necessary for decision makers to sort multiple plans or select the best. In order to improve the accuracy of decision-making, multiple participants are often required to participate in decision-making. The decision-making department can improve decision-making efficiency without affecting the results of the decision. A cut in decision-making cost is often also the main consideration. In the group decision-making process, the decision-maker focuses on the judgment of the problem and the choice of the plan based on his own knowledge and experience, without considering the relationship of other decision-makers, and the decision-maker is giving a decision matrix at the same time. In the process, there is no need to pay too much attention to the relationship between attributes, which is often easy to improve efficiency. This requires the group decision algorithm to calculate the weight of the decision maker and the attribute weight according to the decision matrix given by the decision maker. The decision maker only needs to focus on the problem itself in the whole process. For this reason, for k decision makers, m alternatives, plans with n attributes, and group decision making problems with unknown decision makers' weights and unknown attribute weights can be described in mathematical form. The specific description is given below.

Suppose that a certain decision problem invites k decision makers to give decision data of m alternatives with n attributes, and select the optimal solution among the alternatives based on the data of the decision makers. The attribute weights and the weights of the decision makers are unknown. But the sum of their weights is equal to 1. The set of k decision makers is: $E = \{e_1, e_2, ..., e_k\}$, and the weights satisfy: $\sum_{t=1}^{k} \lambda_t = 1$, $\lambda_t \geq 0 (t = 1, 2, ..., k)$. The set of m schemes is: $Y = \{y_1, y_2, ..., y_m\}$, each scheme contains n identical attributes, the attribute set is: $C = \{C_1, C_2, ..., C_n\}$, and the attribute weight of the weight satisfies: $\sum_{j=1}^{n} \omega_j = 1$, $\omega_j \geq$. For different decision makers, different schemes and attribute values can get a decision matrix. The decision matrix of each decision maker is: $A^{(t)} = (a_{ij}^{(t)})_{m \times n}, (i = 1,2,...m; j = 1,2,...n; t = 1,2,...,k)$, where $a_{ij}^{(t)}$ Represents the decision value of the t-th decision maker for the attribute j in the alternative i, and $a_{ij}^{(k)} = <[u_{a_{ij}}^-, u_{a_{ij}}^+],[v_{a_{ij}}^-, v_{a_{ij}}^+]>^{(t)}$, satisfying that the sum of the qth power of the membership degree and the non-membership q power is less than or equal to 1. The decision matrix of the t-th decision maker is shown in (62).

$$A^{(t)} = \begin{bmatrix} <[u_{a_{11}}^-, u_{a_{11}}^+],[v_{a_{11}}^-, v_{a_{11}}^+]>^{(t)} & \cdots & <[u_{a_{1n}}^-, u_{a_{1n}}^+],[v_{a_{1n}}^-, v_{a_{1n}}^+]>^{(t)} \\ \vdots & \ddots & \vdots \\ <[u_{a_{m1}}^-, u_{a_{m1}}^+],[v_{a_{m1}}^-, v_{a_{m1}}^+]>^{(t)} & \cdots & <[u_{a_{mn}}^-, u_{a_{mn}}^+],[v_{a_{mn}}^-, v_{a_{mn}}^+]>^{(t)} \end{bmatrix} \quad (62)$$

The group decision-making method that integrates IVq-ROFYWA operator and CRITIC-WASPAS given in this paper is used to solve the problem of multi-decision makers and multi-attribute decision-making. In the group decision-making method proposed in this paper, first use the interval value generalized orthogonal ideal distance measurement method to calculate the weight of each decision maker, and then use the decision maker weight and the IVq-ROFYWA operator to aggregate the decision matrix given by each decision maker, and get Assemble the matrix R, and then determine the weight of each attribute through the interval value generalized orthogonal CRITIC method, and then calculate the final interval value score of each program through the interval value generalized orthogonal WASPAS method, and finally determine the ranking and optimal of the alternatives plan.

The decision matrix of aggregating multiple decision makers needs to know the weight of each decision maker, but in actual decision-making situations, the weight of the decision maker is unknown. Therefore, the next section 6.2 will study the need for a decision matrix of multiple decision makers. The process of calculating the weight of decision makers.

## 6.2 Decision maker weight matrix

Koksalmis and Kabak**Error! Reference source not found.** divided decision-maker weight solution methods into five categories: based on similarity, index, clustering, ensemble, and other methods. Among them, literature [70] describes a method based on similarity measurement, which uses ideal distance as a reference point. Find the sum of the distances between each point of each decision matrix and the positive and negative ideal points, and calculate the similarity. After dividing the similarity of each decision maker by the similarity of the decision maker, the weight of each decision maker can be obtained. This method can distinguish each one well. The weight relationship between decision makers. When the decision matrices $A^{(1)}$ and $A^{(2)}$ given by two decision makers are shown in Table 2 and Table 3, it is easy to see that the plan $y_1$ given by the decision maker $e_1$ and the decision The plan $y_1$ given by e_2 has obvious differences, and the plan $y_2$ given by decision maker $e_2$ is also significantly different from the plan $y_2$ given by decision maker $e_2$. The weights of $e_1$ and $e_2$ calculated by reference [70] are respectively : 0.5, 0.5, at this time the weights of the two decision makers are exactly the same. From the matrix given by the decision maker, it is obvious that $e_1$ and $e_2$ have obvious differences, which cannot fully reflect the objective judgment of the decision maker. At the same time, in real life, decision makers are susceptible to their own and external environmental influences when judging different options, such as the psychological state of decision makers, the manifestations of different options, the surrounding environment, and the relationship between group preferences. Therefore, decision makers give different options When judging, obviously showing different weights.

Table 2 Decision matrix $A^{(1)}$

|   | $C_1$ | $C_2$ | $C_3$ |
|---|---|---|---|
| $y_1$ | ([0.02, 0.14],[0.12, 0.93]) | ([0.54, 0.87],[0.02, 0.52]) | ([0.59, 0.82],[0.76, 0.79]) |
| $y_2$ | ([0.13, 0.78],[0.11, 0.61]) | ([0.63, 0.78],[0.01, 0.36]) | ([0.08, 0.2],[0.38, 0.41]) |
| $y_3$ | ([0.70, 0.90],[0.10, 0.20]) | ([0.09, 0.85],[0.65, 0.99]) | ([0.51, 0.87],[0.36, 0.59]) |
| $y_4$ | ([0.01, 0.05],[0.04, 0.94]) | ([0.75, 0.90],[0.53, 0.76]) | ([0.19, 0.31],[0.01, 0.53]) |

Table 3 Decision matrix $A^{(2)}$

|   | $C_1$ | $C_2$ | $C_3$ |
|---|---|---|---|
| $y_1$ | ([0.13, 0.78],[0.11, 0.61]) | ([0.63, 0.78],[0.01, 0.36]) | ([0.08, 0.2],[0.38, 0.41]) |
| $y_2$ | ([0.70, 0.90],[0.10, 0.20]) | ([0.09, 0.85],[0.65, 0.99]) | ([0.51, 0.87],[0.36, 0.59]) |
| $y_3$ | ([0.01, 0.05],[0.04, 0.94]) | ([0.75, 0.90],[0.53, 0.76]) | ([0.19, 0.31],[0.01, 0.53]) |
| $y_4$ | ([0.02, 0.14],[0.12, 0.93]) | ([0.54, 0.87],[0.02, 0.52]) | ([0.59, 0.82],[0.76, 0.79]) |

Inspired by the literature [70], this paper proposes a method for solving the weights of decision makers based on different options. This method uses ideals as a reference point to find the similarity of decision makers to different options, and uses similarity to solve the weights of different options of different decision makers. Thus, the weight matrix of decision makers is obtained, that is, for k decision makers, the weights of m schemes are solved respectively to obtain an m*k decision makers weight matrix. The solution steps are as follows:

（1）Calculate the distance between each element in $A^{(t)}$ and PIS and NIS. That is, using formulas (14) and (15) to calculate the negative ideal distance of $a_{ij}^{(t)}$ $d_{ij}^{-(t)}$ and the positive ideal distance $d_{ij}^{+(t)}$, $(i = 1,2,...,m; j = 1,2,...,n; t = 1,2,...,k)$.

（2）Calculate the similarity between each element in $A^{(t)}$ and the ideal solution $sim_{ij}^{(t)}$, as shown in formula (63), and combine them to form a similarity matrix $SIM^{(t)} = [sim_{ij}^{(t)}]$.

$$sim_{ij}^{(t)} = \frac{d_{ij}^{-(t)}}{d_{ij}^{-(t)} + d_{ij}^{+(t)}} \quad (63)$$

（3）Calculate the sum of the similarity of each scheme, as shown in formula (64)。

$$sim_i^{(t)} = \sum_{j=1}^{n} sim_{ij}^{(t)} \quad (64)$$

（4）Calculate the weight of each plan of each decision maker $\lambda_i^{(t)}$ (t = 1,2,...k; i = 1,2,...m). Use formula (65) to normalize $d_i^{(t)}$ to determine the weight of each plan of each decision maker。

$$\lambda_i^{(t)} = \frac{sim_i^{(t)}}{\sum_{t=1}^{k} sim_i^{(t)}} \quad (65)$$

（5）Solve the decision maker weight matrix $\lambda$. Combine the weights of each scheme calculated in step (4) $\lambda_i^{(t)}(t = 1,2,...k; i = 1,2,...m)$ to form as (66) Shown is the decision-making weight matrix of the decision-maker's scheme. $\lambda_i^{(t)}$ represents the decision weight of the i-th option of the t-th decision maker。

$$\lambda = \begin{bmatrix} \lambda_1^{(1)} & \cdots & \lambda_1^{(k)} \\ \vdots & \ddots & \vdots \\ \lambda_m^{(1)} & \cdots & \lambda_m^{(k)} \end{bmatrix} \quad (66)$$

For the decision matrices $A^{(1)}$ and $A^{(1)}$ given by the decision makers $e_1$ and $e_1$, the method proposed in this paper is used to calculate the weight matrix of the decision makers as shown in Table 4.

Table 4 The weights of decision makers

|  | $e_1(t=1)$ | $e_2(t=2)$ |
|---|---|---|
| $\lambda_1^{(t)}$ | 0.46 | 0.54 |
| $\lambda_2^{(t)}$ | 0.49 | 0.51 |
| $\lambda_3^{(t)}$ | 0.55 | 0.45 |
| $\lambda_4^{(t)}$ | 0.49 | 0.51 |

It can be seen from Table 4 that the weights of different schemes of different decision makers are obviously inconsistent. On the one hand, it reflects the real judgment state of the decision makers, and on the other hand, it also reflects the reaction of the decision makers when judging different schemes. Literature [70] gives a weight for the decision maker to judge the whole problem, which cannot reflect the change of the decision maker due to the external environment. However, the solution method proposed in this paper is that different solutions have a weight. The weight is unknown, and the decision maker When there are obvious differences in the judgment of different schemes, it is more able to reflect the real situation.

## 6.3 Group decision method

This section combines the interval-valued generalized orthogonal fuzzy Yager weighted average operator (IVq-ROFYWA) in Section 3.2 and the interval-valued generalized orthogonal fuzzy CRTIC-WASPAS method in Section 4.3 to select the optimal solution, and studies the integration of IVq-ROFYWA. The specific operation process of the interval-valued generalized orthogonal fuzzy group decision-making method of CRITIC-WASPAS fusion method solves the decision-making problem of multiple decision-makers and multiple attributes.

The specific processing of the interval-valued generalized orthogonal fuzzy group decision-making method integrating the Yager aggregation operator and the CRITIC-WASPAS method includes the following steps:

(1)Determine the value of q. According to the decision matrix given by each decision maker, select the q value that satisfies the interval generalized orthogonal condition. That is, according to $(u_{a_{ij}^{(t)}}^+)^q + (v_{a_{ij}^{(t)}}^+)^q \leq 1 (t = 1,2,...k)$, select the appropriate q.

(2)Calculate the weight of each decision maker. According to the interval-based generalized orthogonal negative ideal distance measurement method in Section 6.2, the decision weight $\lambda_t(t = 1,2,...k)$ of each decision maker can be obtained.

(3)Assemble the decision matrix of k decision makers. Using interval value generalized orthogonal fuzzy Yager weighted average IVq-ROFYWA operator, the decision matrix of k decision makers is assembled $A = (A^{(1)}, A^{(2)}, A^{(3)},..., A^{(k)})$, get the aggregation matrix $R = [r_{ij}]_{m \times n}$.

The calculation of the aggregation matrix is shown in formula (67) ($i = 1,2,...,m$ ;$j = 1,2,...,n; t = 1,2,...,k$).

$$r_{ij} = IVq - ROFYWA(a_{ij}^{(1)}, a_{ij}^{(2)}, a_{ij}^{(3)},...,a_{ij}^{(t)}); \quad (67)$$

（4）Calculate the interval value attribute weight of the matrix. First, the interval-valued attribute weight $w_j = <[u_{w_j}^-, u_{w_j}^+],[v_{w_j}^-, v_{w_j}^+]>$ is obtained by the interval-valued generalized orthogonal fuzzy CRITIC method.

(5) Calculate the real number attribute weight of the matrix. Using the interval value generalized orthogonal ideal distance measurement method, the real number weight $\omega_j (j = 1,2,...n)$ of each attribute is determined by formula (61).

(6) Calculate the interval evaluation value of the scheme. According to the real number weight of each attribute determined in step (5), the interval value generalized orthogonal fuzzy WASPAS method is used to select the appropriate λ to obtain the interval evaluation value $r_i = <[u_{r_i}^-, u_{r_i}^+],[v_{r_i}^-, v_{r_i}^+]> (i = 1,2,...m)$.

(7) Calculate the score function value of each scheme. The interval value generalized orthogonal score function (formula 20) is used to calculate the score function value $S(r_i)$ ($i = 1,2,...m$) of each scheme, and the score function value is a real number at this time.

(8) Determine the order of the schemes. According to the score function value of each scheme calculated in step (7), the score function value of each scheme is compared to determine the order of the schemes.

(9) Choose the best plan. According to the order of the schemes in step (8), the optimal scheme is selected.

The processing flow chart of the interval-valued generalized orthogonal fuzzy group decision-making method integrating the Yager aggregation operator and the CRITIC-WASPAS method is shown in Figure 2:

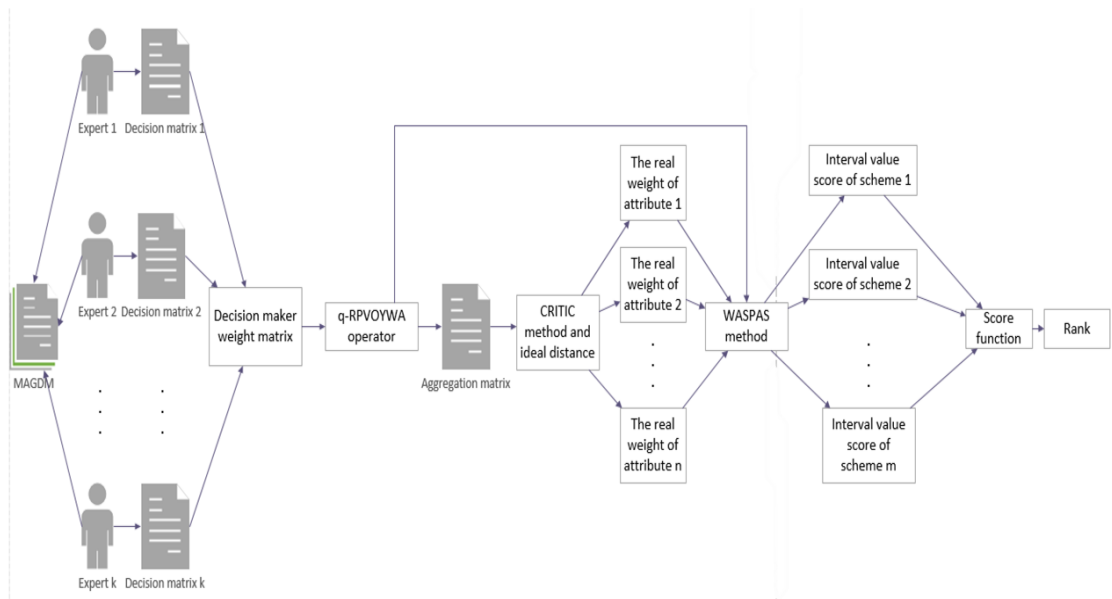

Figure 2 The processing flow chart of the group decision method

The figure briefly describes the processing flow of the entire group decision-making method: collect the decision matrices of k decision makers, aggregate the decision matrices of k decision makers through the IVq-ROFYWA operator, and obtain an aggregate matrix, using the CRITIC method and negative ideal distance The measurement method calculates the real weight of each attribute, inputs the attribute weight and aggregation matrix into the WSASPAS method to obtain the interval value score of each scheme, and finally determines the order of the schemes by the score function.

# 7. Application case

This section is divided into 4 subsections to verify the methods and models proposed in this article. The case of hypertension detection and prediction management system described in section 7.1 confirms the group decision-making method proposed in section 6.3 of this article. Section 7.2 analyzes the impact of group decision-making method ranking and verification negative The advantages of ideal distance calculation weights. Sections 7.3 and 7.4 respectively compare and study the operators and decision-making methods in this paper with other operators and decision-making methods, which further prove the feasibility and effectiveness of the group decision-making method proposed in this paper.

## 7.1 Case

With the development of information technology, it was applied to hypertension prevention and treatment for community management. Although the existing management system satisfies the basis of medical management, the efficiency needs to be improved. To develop a community hypertension management system, in order to provide more accurate monitoring and prediction of hypertension susceptible populations, the prediction model that the system needs to build after using big data processing still needs to be analyzed and judged by decision makers. Residents' susceptibility to hypertension is divided into five levels: standard, not susceptible, at risk of disease, high risk of disease, and very high risk of disease in recent years. The hypertension management system uses green, blue, yellow, and Orange and red are indicated, and they are divided into five options $(y_1,y_2,y_3,y_4,y_5)$, each color and the corresponding risk level are shown in Table 5. For each plan, the warning level is related to many factors of hypertensive patients. Considering that the collected data information is mainly physical examination information, the experts judge whether residents are vulnerable to risk mainly depends on: current blood pressure measurement value, genetic history, The five indicators of changing trend of multiple blood pressure measurements, age, and related risk factors are represented by C1, C2, C3, C4, and C5. All indicators are efficiency indicators.

Table 5 The meaning of the plan early warning

| plan | Risk level | Warning color |
| --- | --- | --- |
| $y_1$ | Very high risk of illness in recent years | Red |
| $y_2$ | High risk of illness | Orange |
| $y_3$ | At risk of illness | Yellow |
| $y_4$ | Not susceptible | Blue |
| $y_5$ | Normal | Green |

During the system design process, 5 hypertension experts $(e_1,e_2,e_3,e_4,e_5)$ were hired to analyze the physical examination data. In the early stage of system implementation, experts judged the residents' blood pressure measurement value, the change trend of multiple blood pressure measurement values, age, genetic history, and related risk factors based on their own experience.

Because predicted hypertension risk is classified into multiple ranges, instead of assigned a certain value, the judgment value given by the expert is expressed as an interval value. The existing 1 resident is 50 years old, systolic blood pressure: 135mmHg, diastolic blood pressure: 87mmHg, blood pressure has been rising in the past 5 years, there is a genetic history, no related diseases, normal living habits, p=2 in IVq-ROFYWA, 5 high The interval values of blood pressure experts to judge the risk of hypertension for the residents are shown in Table 6 to Table 10:

Table 6 Decision matrix given by Expert 1 $A^{(1)}$

|  | C1 | C2 | C3 | C4 | C5 |
|---|---|---|---|---|---|
| $y_1$ | ([0.85,0.95],[0.1,0.2]) | ([0.8,0.9],[0.1,0.2]) | ([0.85,0.95], [0.1,0.2]) | ([0.7,0.8],[0.2,0.3]) | ([0.4,0.5],[0.5,0.6]) |
| $y_2$ | ([0.9,0.95],[0.1,0.2]) | ([0.75,0.8],[0.2,0.3]) | ([0.85,0.9], [0.1,0.15]) | ([0.65,0.7],[0.3,0.35]) | ([0.5,0.6],[0.4,0.5]) |
| $y_3$ | ([0.8,0.9],[0.1,0.2]) | ([0.6,0.7],[0.3,0.4]) | ([0.8,0.95], [0.1,0.2]) | ([0.6,0.7],[0.3,0.4]) | ([0.3,0.4],[0.6,0.7]) |
| $y_4$ | ([0.5,0.6],[0.4,0.5]) | ([0.6,0.75],[0.2,0.3]) | ([0.7,0.8], [0.2,0.3]) | ([0.5,0.6],[0.4,0.5]) | ([0.3,0.4],[0.6,0.7]) |
| $y_5$ | ([0.4,0.5],[0.5,0.6]) | ([0.5,0.6],[0.4,0.5]) | ([0.6,0.7], [0.3,0.4]) | ([0.5,0.6],[0.4,0.5]) | ([0.1,0.2],[0.8,0.9]) |

Table 7 Decision matrix given by Expert 2 $A^{(2)}$

|  | C1 | C2 | C3 | C4 | C5 |
|---|---|---|---|---|---|
| $y_1$ | ([0.7,0.8],[0.2,0.3]) | ([0.8,0.9],[0.1,0.2]) | ([0.8,0.9], [0.1,0.2]) | ([0.65,0.75],[0.3,0.35]) | ([0.5,0.6],[0.4,0.5]) |
| $y_2$ | ([0.9,0.95],[0.1,0.2]) | ([0.75,0.85],[0.2,0.3]) | ([0.85,0.9], [0.1,0.15]) | ([0.6,0.7],[0.3,0.4]) | ([0.4,0.5],[0.5,0.6]) |
| $y_3$ | ([0.8,0.9],[0.1,0.2]) | ([0.65,0.7],[0.2,0.3]) | ([0.75,0.8], [0.2,0.3]) | ([0.7,0.8],[0.2,0.3]) | ([0.35,0.5],[0.5,0.65]) |
| $y_4$ | ([0.7,0.8],[0.2,0.3]) | ([0.6,0.7],[0.3,0.4]) | ([0.7,0.8], [0.2,0.3]) | ([0.2,0.3],[0.7,0.8]) | ([0.2,0.3],[0.7,0.8]) |
| $y_5$ | ([0.75,0.85],[0.2,0.3]) | ([0.55,0.65],[0.35,0.4]) | ([0.6,0.7], [0.3,0.4]) | ([0.1,0.2],[0.8,0.9]) | ([0.3,0.4],[0.6,0.7]) |

Table 8 Decision matrix given by Expert 3 $A^{(3)}$

|  | C1 | C2 | C3 | C4 | C5 |
|---|---|---|---|---|---|
| $y_1$ | ([0.85,0.9],[0.1,0.2]) | ([0.7,0.8],[0.2,0.3]) | ([0.85,0.9], [0.1,0.2]) | ([0.65,0.75],[0.25,0.35]) | ([0.4,0.5],[0.5,0.6]) |
| $y_2$ | ([0.9,0.95],[0.1,0.2]) | ([0.8,0.9],[0.1,0.2]) | ([0.85,0.95], [0.1,0.15]) | ([0.65,0.8],[0.2,0.3]) | ([0.5,0.6],[0.4,0.5]) |
| $y_3$ | ([0.85,0.9],[0.1,0.2]) | ([0.6,0.7],[0.3,0.4]) | ([0.75,0.85], [0.2,0.3]) | ([0.6,0.7],[0.3,0.4]) | ([0.4,0.5],[0.5,0.6]) |
| $y_4$ | ([0.85,0.9],[0.1,0.2]) | ([0.4,0.5],[0.5,0.6]) | ([0.2,0.3], [0.7,0.8]) | ([0.4,0.5],[0.5,0.6]) | ([0.3,0.4],[0.6,0.7]) |
| $y_5$ | ([0.7,0.85],[0.2,0.3]) | ([0.2,0.3],[0.8,0.9]) | ([0.1,0.2], [0.8,0.9]) | ([0.3,0.4],[0.6,0.7]) | ([0.3,0.4],[0.6,0.7]) |

Table 9 Decision matrix given by Expert 4 $A^{(4)}$

|  | C1 | C2 | C3 | C4 | C5 |
|---|---|---|---|---|---|
| $y_1$ | ([0.85,0.9],[0.1,0.2]) | ([0.7,0.8],[0.2,0.3]) | ([0.85,0.95], [0.1,0.15]) | ([0.6,0.75],[0.3,0.35]) | ([0.45,0.5],[0.5,0.55]) |
| $y_2$ | ([0.9,0.95],[0.1,0.2]) | ([0.75,0.8],[0.2,0.3]) | ([0.85,0.95], [0.1,0.15]) | ([0.7,0.75],[0.2,0.3]) | ([0.4,0.5],[0.5,0.6]) |
| y₃ | ([0.75,0.8],[0.2,0.3]) | ([0.6,0.75],[0.3,0.4]) | ([0.8,0.9], [0.1,0.2]) | ([0.6,0.7],[0.3,0.4]) | ([0.35,0.45],[0.5,0.65]) |
| y₄ | ([0.85,0.9],[0.1,0.2]) | ([0.4,0.5],[0.5,0.6]) | ([0.5,0.6], [0.4,0.5]) | ([0.6,0.7],[0.3,0.4]) | ([0.4,0.5],[0.5,0.6]) |
| y₅ | ([0.7,0.85],[0.2,0.3]) | ([0.3,0.4],[0.6,0.7]) | ([0.4,0.5], [0.5,0.6]) | ([0.45,0.5],[0.5,0.6]) | ([0.35,0.4],[0.5,0.6]) |

Table 10 Decision matrix given by expert 5 $A^{(5)}$

| | C1 | C2 | C3 | C4 | C5 |
|---|---|---|---|---|---|
| y₁ | ([0.8,0.9],[0.1,0.2]) | ([0.7,0.8],[0.2,0.3]) | ([0.8,0.9], [0.1,0.2]) | ([0.6,0.75],[0.3,0.35]) | ([0.4,0.5],[0.5,0.6]) |
| y₂ | ([0.9,0.95],[0.1,0.2]) | ([0.75,0.85],[0.2,0.3]) | ([0.8,0.95], [0.1,0.2]) | ([0.7,0.8],[0.2,0.3]) | ([0.5,0.6],[0.4,0.5]) |
| y₃ | ([0.7,0.85],[0.1,0.2]) | ([0.65,0.75],[0.3,0.4]) | ([0.8,0.9], [0.1,0.2]) | ([0.6,0.7],[0.2,0.3]) | ([0.4,0.5],[0.5,0.6]) |
| y₄ | ([0.8,0.9],[0.1,0.2]) | ([0.5,0.6],[0.4,0.5]) | ([0.6,0.7], [0.3,0.4]) | ([0.6,0.65],[0.3,0.4]) | ([0.35,0.45],[0.5,0.6]) |
| y₅ | ([0.75,0.85],[0.2,0.3]) | ([0.4,0.5],[0.5,0.6]) | ([0.5,0.6], [0.4,0.5]) | ([0.3,0.4],[0.6,0.7]) | ([0.35,0.4],[0.5,0.6]) |

According to the known conditions, this paper adopts the interval value generalized orthogonal fuzzy group decision-making method integrating IVq-ROFYWA operator and CRITIC-WASPAS method to select the early warning risk plan, which is used for the monitoring and prediction of hypertension susceptible population. Since all indicators are efficiency indicators, the attributes are all positive attributes. The entire decision-making process includes the following 10 steps:

(1) Determine the appropriate q value. It can be observed from Table 7 to Table 11 that when the value of q is 3, the condition of generalized orthogonal fuzzy number is satisfied: $(u_{a_{ij}^{(t)}}^+)^q + (v_{a_{ij}^{(t)}}^+)^q \leq 1$;

(2) Determine the weight matrix $\lambda$ of each expert. The expert's decision weight matrix can be determined by formulas (63) to (66), and the subdivision steps are as follows:

(2.1) The similarity matrix $SIM^{(t)}$ of each expert is shown below, and the result is rounded to 2 decimal places:

$$SIM^{(1)} = \begin{bmatrix} 0.87 & 0.81 & 0.87 & 0.71 & 0.46 \\ 0.89 & 0.72 & 0.83 & 0.64 & 0.54 \\ 0.81 & 0.62 & 0.84 & 0.62 & 0.38 \\ 0.54 & 0.65 & 0.71 & 0.54 & 0.38 \\ 0.46 & 0.54 & 0.62 & 0.54 & 0.19 \end{bmatrix}$$

$$SIM^{(2)} = \begin{bmatrix} 0.71 & 0.81 & 0.81 & 0.66 & 0.54 \\ 0.89 & 0.75 & 0.83 & 0.62 & 0.46 \\ 0.81 & 0.65 & 0.72 & 0.71 & 0.44 \\ 0.71 & 0.62 & 0.71 & 0.29 & 0.29 \\ 0.54 & 0.58 & 0.62 & 0.19 & 0.38 \end{bmatrix}$$

$$SIM^{(3)} = \begin{bmatrix} 0.83 & 0.71 & 0.83 & 0.66 & 0.46 \\ 0.89 & 0.81 & 0.87 & 0.69 & 0.54 \\ 0.83 & 0.62 & 0.75 & 0.62 & 0.46 \\ 0.83 & 0.46 & 0.29 & 0.46 & 0.38 \\ 0.73 & 0.20 & 0.19 & 0.38 & 0.38 \end{bmatrix}$$

$$SIM^{(4)} = \begin{bmatrix} 0.83 & 0.71 & 0.87 & 0.64 & 0.48 \\ 0.89 & 0.72 & 0.87 & 0.68 & 0.46 \\ 0.72 & 0.64 & 0.81 & 0.62 & 0.43 \\ 0.83 & 0.46 & 0.54 & 0.62 & 0.46 \\ 0.73 & 0.38 & 0.46 & 0.47 & 0.44 \end{bmatrix}$$

$$SIM^{(5)} = \begin{bmatrix} 0.81 & 0.71 & 0.81 & 0.64 & 0.46 \\ 0.89 & 0.75 & 0.84 & 0.71 & 0.54 \\ 0.74 & 0.65 & 0.81 & 0.63 & 0.46 \\ 0.81 & 0.54 & 0.62 & 0.60 & 0.45 \\ 0.75 & 0.46 & 0.54 & 0.38 & 0.44 \end{bmatrix}$$

(2.2) The sum $sim_i$ of the similarity of each plan of each expert is shown in Table 11. The result is to 2 decimal places:

Table 11 The sum $sim_i$ of the similarity of each plan of each expert

|  | $e_1$ | $e_2$ | $e_3$ | $e_4$ | $e_5$ |
|---|---|---|---|---|---|
| $y_1$ | 3.72 | 3.53 | 3.49 | 3.53 | 3.43 |
| $y_2$ | 3.62 | 3.55 | 3.80 | 3.62 | 3.73 |
| $y_3$ | 3.27 | 3.33 | 3.28 | 3.22 | 3.29 |
| $y_4$ | 2.82 | 2.62 | 2.42 | 2.91 | 3.02 |
| $y_5$ | 2.35 | 2.31 | 1.88 | 2.48 | 2.57 |

(2.3) Normalize the sum of the similarity of each plan of each expert, and obtain the weight matrix $\lambda$ of each expert as shown below, and the result retains 4 decimal places:

$$\lambda = \begin{bmatrix} 0.2102 & 0.1994 & 0.1972 & 0.1994 & 0.1938 \\ 0.1976 & 0.1938 & 0.2074 & 0.1976 & 0.2036 \\ 0.1995 & 0.2032 & 0.2001 & 0.1965 & 0.2007 \\ 0.2045 & 0.1900 & 0.1755 & 0.2110 & 0.2190 \\ 0.2028 & 0.1993 & 0.1622 & 0.2140 & 0.2217 \end{bmatrix}$$

(3) Using the IVq-ROFYWA operator, use the formula (67) in Section 6.3 to aggregate the decision matrices of the five experts to obtain the aggregate matrix $R = [r_{ij}]$, as shown in Table 12, and the result retains 2 decimal places:

Table 12 Assembly matrix $R$

|  | C1 | C2 | C3 | C4 | C5 |
|---|---|---|---|---|---|
| $y_1$ | ([0.82,0.9],[0.13,0.23]) | ([0.75,0.85],[0.17,0.27]) | ([0.83,0.92], [0.1,0.19]) | ([0.65,0.76],[0.27,0.34]) | ([0.44,0.53],[0.48,0.57]) |
| $y_2$ | ([0.9,0.95],[0.1,0.2]) | ([0.76,0.84],[0.19,0.28]) | ([0.84,0.93], [0.1,0.16]) | ([0.67,0.76],[0.25,0.33]) | ([0.47,0.57],[0.44,0.54]) |
| $y_3$ | ([0.79,0.87],[0.13,0.23]) | ([0.62,0.72],[0.28,0.38]) | ([0.78,0.89], [0.16,0.25]) | ([0.63,0.73],[0.27,0.37]) | ([0.37,0.48],[0.52,0.64]) |
| $y_4$ | ([0.78,0.85],[0.25,0.33]) | ([0.53,0.65],[0.41,0.50]) | ([0.63,0.72], [0.42,0.50]) | ([0.54,0.61],[0.47,0.56]) | ([0.34,0.44],[0.58,0.68]) |
| $y_5$ | ([0.67,0.79],[0.35,0.43]) | ([0.46,0.55],[0.55,0.63]) | ([0.53,0.63], [0.49,0.57]) | ([0.42,0.49],[0.59,0.68]) | ([0.32,0.39],[0.61,0.70]) |

(4) Use the interval value generalized fuzzy CRITIC method in Section 5.2 to calculate the attribute weights of the aggregation matrix R, which includes multiple subdivision steps:

(4.1) The standardized decision matrix that can be obtained from formula (50) is shown in Table 13, and the result is rounded to 2 decimal places:

Table 13 Standardized Decision Matrix X

|  | C1 | C2 | C3 | C4 | C5 |
|---|---|---|---|---|---|
| $y_1$ | ([0.68,0.81],[0.21,0.32]) | ([0.55,0.68],[0.19,0.3]) | ([0.59,0.71], [0.22,0.34]) | ([0.52,0.64],[0.18,0.28]) | ([0.51,0.63],[0.15,0.25]) |
| y₂ | ([0.69,0.81],[0.21,0.32]) | ([0.56,0.67],[0.19,0.3]) | ([0.59,0.71], [0.22,0.34]) | ([0.53,0.64],[0.18,0.27]) | ([0.51,0.64],[0.14,0.24]) |
| y₃ | ([0.68,0.81],[0.21,0.32]) | ([0.52,0.64],[0.2,0.32]) | ([0.58,0.71], [0.22,0.34]) | ([0.51,0.63],[0.18,0.28]) | ([0.5,0.62],[0.16,0.27]) |
| y₄ | ([0.68,0.8],[0.21,0.33]) | ([0.5,0.62],[0.22,0.35]) | ([0.56,0.67], [0.25,0.37]) | ([0.49,0.59],[0.22,0.34]) | ([0.5,0.61],[0.17,0.28]) |
| y₅ | ([0.68,0.8],[0.22,0.33]) | ([0.49,0.6],[0.26,0.39]) | ([0.55,0.66], [0.26,0.39]) | ([0.47,0.57],[0.25,0.38]) | ([0.5,0.61],[0.18,0.29]) |

(4.2) The average value of each column obtained by formula (54), and the result is rounded to 2 decimal places:

$$\bar{x}_{C_1} =< [0.68,0.81],[0.21,0.32] > \quad \bar{x}_{C_2} =< [0.52,0.65],[0.22,0.34] >$$

$$\bar{x}_{C_3} =< [0.58,0.69],[0.24,0.36] > \quad \bar{x}_{C_4} =< [0.51,0.62],[0.21,0.32] >$$

$$\bar{x}_{C_5} =< [0.50,0.62],[0.16,0.27] >$$

(4.3) The correlation coefficients between the attributes that can be obtained from the formula (53) are shown in Table 14, and the results are rounded to 2 decimal places:

Table 14 Correlation coefficients between attributes

| | C1 | C2 | C3 | C4 (age) | C5 |
|---|---|---|---|---|---|
| $C_1$ | ([0.81,0.81],[0.03,0.11]) | ([0.75,0.75],[0.02,0.09]) | ([0.77,0.77],[0.03,0.1]) | ([0.75,0.75],[0.02,0.08]) | ([0.74,0.74],0.01,0.06]) |
| $C_2$ | ([0.75,0.75],[0.02,0.09]) | ([0.66,0.66],[0.02,0.07]) | ([0.69,0.69],[0.02,0.08]) | ([0.65,0.65],[0.01,0.06]) | ([0.64,0.64],[0.01,0.05]) |
| $C_3$ | ([0.77,0.77],[0.03,0.1]) | ([0.69,0.69],[0.02,0.08]) | ([0.72,0.72], [0.02,0.09]) | ([0.69,0.69],[0.02,0.07]) | ([0.68,0.68], [0.01,0.06]) |
| $C_4$ | ([0.75,0.75],[0.02,0.08]) | ([0.65,0.65],[0.01,0.06]) | ([0.69,0.69],[0.02,0.07]) | ([0.65,0.65],[0.01,0.05]) | ([0.64,0.64],[0.01,0.04]) |
| $C_5$ | ([0.74,0.74],[0.01,0.06]) | ([0.64,0.64],[0.01,0.05]) | ([0.68,0.68], [0.01,0.06]) | ([0.64,0.64],[0.01,0.04]) | ([0.62,0.62],[0.01,0.04]) |

(4.4) The standard deviation of each column obtained by formula (55), and the result is rounded to 2 decimal places:

$$\sigma_{C_1} =< [0.82,0.69],[0.19,0.34] > \quad \sigma_{C_2} =< [0.68,0.54],[0.15,0.29] >$$

$$\sigma_{C_3} =< [0.72,0.59],[0.18,0.33] > \quad \sigma_{C_4} =< [0.65,0.53],[0.14,0.26] >$$

$$\sigma_{C_5} =< [0.64,0.51],[0.11,0.22] >$$

(4.5) The index value of each attribute obtained by formula (56), and the result is kept to 3 decimal places:

$$N_{C_1} =< [0.034,0.108],[0.289,0.383] > \quad N_{C_2} =< [0.020,0.067],[0.184,0.301] >$$

$$N_{C_3} =< [0.027,0.083],[0.226,0.346] > \quad N_{C_4} =< [0.017,0.057],[0.176,0.273] >$$

$$N_{C_5} =< [0.011,0.045],[0.151,0.233] >$$

(4.6) The index proportion of each attribute obtained by formula (57), that is, the weight of the interval value, the result is kept to 3 decimal places:

$$w_{C_1} =< [0.289,0.386],[0.010,0.041] > \quad w_{C_2} =< [0.184,0.302],[0.004,0.020] >$$

$$w_{C_3} =< [0.226,0.348],[0.006,0.029] > \quad w_{C_4} =< [0.176,0.274],[0.003,0.016] >$$

$$w_{C_5} =< [0.151,0.234],[0.002,0.010] >$$

According to the 6 subdivision steps of the CRITIC method, the interval value weight $w_{C_i}$ (i = 1,2,3,4,5) of each attribute can be determined.

(5) Calculate the negative ideal distance $w_{C_i}(i = 1,2,3,4,5)$ of the interval value weight $d_{C_i}$ $(i = 1,2,3,4,5)$ of each indicator as shown below, and the result shall be 3 decimal places:

$$d_{C_1} = 0.520 \quad d_{C_2} = 0.508 \quad d_{C_3} = 0.513 \quad d_{C_4} = 0.507 \quad d_{C_5} = 0.504$$

(6) According to the negative ideal distance $d_{C_i}(i = 1,2,3,4,5)$ obtained in step (5), the real number weights $\omega_j(j = 1,2,3,4,5)$ of 5 indicators can be determined by formula (61). The real number weights of 5 indicators are as follows, and the result retains 5 decimal places:

$$\omega_1 = 0.20376 \quad \omega_2 = 0.19906 \quad \omega_3 = 0.20102 \quad \omega_4 = 0.19867 \quad \omega_5 = 0.19749$$

(7) According to the real number weights $\omega_j(j = 1,2,3,4,5)$ of the five indicators obtained in (6), the interval value generalized fuzzy WASPAS method of section 5.3 is used to calculate the interval value score of the scheme of the aggregation matrix R, which includes multiple subdivision steps:

(7.1) The standardized decision matrix obtained by formula (50) is shown in Table 15, and the result is rounded to 2 decimal places:

Table 15 Standardized decision matrix R*

|  | C1 | C2 | C3 (change trend) | C4 (age) | C5 |
|---|---|---|---|---|---|
| $y_1$ | ([0.68,0.81],[0.21,0.32]) | ([0.55,0.68],[0.19,0.3]) | ([0.59,0.71], [0.22,0.34]) | ([0.52,0.64],[0.18,0.28]) | ([0.51,0.63],[0.15,0.25]) |
| $y_2$ | ([0.69,0.81],[0.21,0.32]) | ([0.56,0.67],[0.19,0.3]) | ([0.59,0.71], [0.22,0.34]) | ([0.53,0.64],[0.18,0.27]) | ([0.51,0.64],[0.14,0.24]) |
| $y_3$ | ([0.68,0.81],[0.21,0.32]) | ([0.52,0.64],[0.2,0.32]) | ([0.58,0.71], [0.22,0.34]) | ([0.51,0.63],[0.18,0.28]) | ([0.5,0.62],[0.16,0.27]) |
| $y_4$ | ([0.68,0.8],[0.21,0.33]) | ([0.5,0.62],[0.22,0.35]) | ([0.56,0.67], [0.25,0.37]) | ([0.49,0.59],[0.22,0.34]) | ([0.5,0.61],[0.17,0.28]) |
| $y_5$ | ([0.68,0.8],[0.22,0.33]) | ([0.49,0.6],[0.26,0.39]) | ([0.55,0.66], [0.26,0.39]) | ([0.47,0.57],[0.25,0.38]) | ([0.5,0.61],[0.18,0.29]) |

(7.2) The weighted sum of each scheme is obtained by formula (58), and the result is kept to 3 decimal places:

$$Q^{(1)}{}_{y_1} = <[0.588,0.711],[0.193,0.301]> \quad Q^{(1)}{}_{y_2} = <[0.594,0.710],[0.192,0.298]>$$

$$Q^{(1)}{}_{y_3} = <[0.581,0.703],[0.196,0.308]> \quad Q^{(1)}{}_{y_4} = <[0.573,0.683],[0.217,0.337]>$$

$$Q^{(1)}{}_{y_5} = <[0.568,0.677],[0.238,0.360]>$$

(7.3) The weighted product of each scheme is obtained by formula (59), and the result is kept to 3 decimal places:

$$Q^{(1)}{}_{y_1} = <[0.575,0.695],[0.197,0.306]> \quad Q^{(1)}{}_{y_2} = <[0.579,0.695],[0.197,0.304]>$$
$$Q^{(1)}{}_{y_3} = <[0.564,0.684],[0.200,0.312]> \quad Q^{(1)}{}_{y_4} = <[0.553,0.661],[0.221,0.340]>$$
$$Q^{(1)}{}_{y_5} = <[0.546,0.652],[0.242,0.366]>$$

(7.4) The final interval evaluation value of each scheme obtained by formula (60), $\lambda = 0.5$, the result shall be 6 decimal places:

$$r_1 = <[0.588,0.703],[0.195,0.303]> \quad r_2 = <[0.587,0.703],[0.194,0.301]>$$

$$r_3 = <[0.573,0.694],[0.198,0.310]> \quad r_4 = <[0.563,0.672],[0.219,0.338]>$$
$$r_5 = <[0.557,0.665],[0.240,0.363]>$$

According to the 4 subdivision steps of the above WASPAS method, the interval value score of each scheme can be calculated $r_i(i = 1,2,3,4,5)$。

(8) Using the score function proposed in section 3.2, the values of $\alpha$ and $\beta$ that have been

changed many times are used to calculate $r_i$(i = 1,2,3,4,5), Here we assume that both $\alpha$ and $\beta$ are 0.5. The calculation result of the score function is shown below, and the result is kept 5. Decimals:

$$S(r_1) = 0.88524 \quad S(r_2) = 0.88615 \quad S(r_3) = 0.88217 \quad S(r_4) = 0.87352 \quad S(r_5) = 0.86716$$

Comparing the size of the score value of each scheme can be obtained: $S(r_2) > S(r_1) > S(r_3) > S(r_4) > S(r_5)$。

(9) According to the comparison result of step (8), the order of the schemes can be obtained: $y_2 > y_1 > y_3 > y_4 > y_5$。

According to the calculation results of the group decision method proposed in this article, it can be found that it is the optimal solution, so the hypertension risk management system's early warning of the residents' future risk of hypertension is an orange warning. Therefore, the patient has a high risk of developing diseases in the future and needs to consult a doctor on time to prevent hypertension in time. This result is consistent with the judgment of 5 hypertension experts. It can be concluded that the group decision-making method proposed in this paper is feasible and effective.

### 7.2.1 Case analysis

### 7.2.1 Case Method Analysis

In order to further verify the feasibility of the group decision-making method proposed in this paper, this section mainly studies the influence of the variables in the group decision-making method on the ranking of alternatives and comparatively analyzes the advantages of negative ideal distance in the case.

To verify the influence of the change of q in the interval value generalized orthogonal fuzzy environment on the group decision-making method in this paper, the text takes the IVq-ROFYWA operator aggregation as an example. When p=2, q takes 2, 3, 4, and 5, multiple times Calculate the changed values of $\alpha$ and $\beta$, and obtain the score ranking of each plan and the ranking result of the plan as shown in Table 16. It can be seen from the table that the ranking result of the plan does not change with the increase of q. At the same time, when p=2, q changes, $\alpha$ and $\beta$ When both are 0.5, the order and trend of the schemes are shown in Figure 3.

Table 16 Ranking of schemes under different q values

| q | Score sort | Scheme ranking | Result |
|---|---|---|---|
| 2 | $Sr_2>Sr_1>Sr_3>Sr_4>Sr_5$ | $y_2>y_1>y_3>y_4>y_5$ | Orange warning |
| 3 | $Sr_2>Sr_1>Sr_3>Sr_4>Sr_5$ | $y_2>y_1>y_3>y_4>y_5$ | Orange warning |
| 4 | $Sr_2>Sr_1>Sr_3>Sr_4>Sr_5$ | $y_2>y_1>y_3>y_4>y_5$ | Orange warning |
| 5 | $Sr_2>Sr_1>Sr_3>Sr_4>Sr_5$ | $y_2>y_1>y_3>y_4>y_5$ | Orange warning |

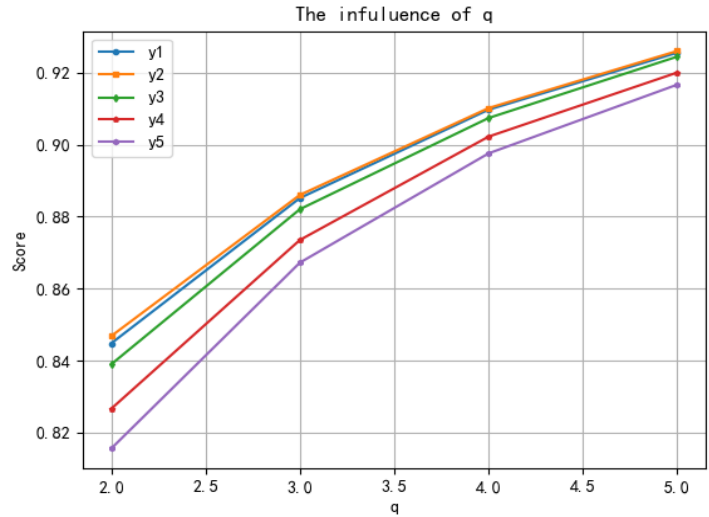

Figure 3 Scheme ranking results under different q values

In Figure 3, when q=4, $S(r_1) = 0.90978, S(r_2) = 0.91013, S(r_1) < S(r_2)$; when q=5, $S(r_1) = 0.92569, S(r_2) = 0.9261, S(r_1) < S(r_2)$。 It can be seen from the results in Figure 3 that when q is not the same, the order of the solutions does not change.

From the formula of the IVq-ROFYWA operator, we can see that there is also p that affects the aggregation. So we further study the influence of the change of p under the interval-value generalized orthogonal fuzzy environment on the group decision-making method in this paper. When q=3 and p takes 1, 2, 3, 4, and 5 respectively, calculate the changed values of $\alpha$ and $\beta$ multiple times, The score ranking and the ranking results of the schemes are shown in Table 17. From the table, it can be seen that the ranking results of the schemes do not change with the increase of p. At the same time, when q=3, p changes, and $\alpha$ and $\beta$ are both 0.5, the calculation result of the score function of the scheme is shown in Figure 4. From the results, it can be seen that when p is not the same, the order of the scheme does not change.

Table 17 The ranking of schemes under different p values

| p | Score sort | Scheme ranking | Result |
|---|---|---|---|
| 1 | $Sr_2 > Sr_1 > Sr_3 > Sr_4 > Sr_5$ | $y_2 > y_1 > y_3 > y_4 > y_5$ | Orange warning |
| 2 | $Sr_2 > Sr_1 > Sr_3 > Sr_4 > Sr_5$ | $y_2 > y_1 > y_3 > y_4 > y_5$ | Orange warning |
| 3 | $Sr_2 > Sr_1 > Sr_3 > Sr_4 > Sr_5$ | $y_2 > y_1 > y_3 > y_4 > y_5$ | Orange warning |
| 4 | $Sr_2 > Sr_1 > Sr_3 > Sr_4 > Sr_5$ | $y_2 > y_1 > y_3 > y_4 > y_5$ | Orange warning |
| 5 | $Sr_2 > Sr_1 > Sr_3 > Sr_4 > Sr_5$ | $y_2 > y_1 > y_3 > y_4 > y_5$ | Orange warning |

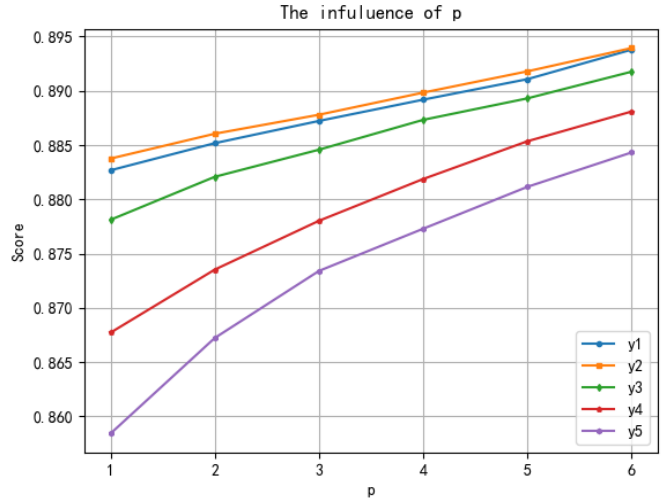

Figure 4 Scheme ranking results under different p values

Continue to study the impact of simultaneous changes in p and q on the order of the schemes. When p=2, 3, 4, 5; q=3, 4, 5, 6, 7, and calculate the changed values of $\alpha$ and $\beta$ multiple times, the scheme The result of the score function is shown in Table 18. It can be seen that when p and q change at the same time, the ranking of the scheme remains unchanged. At the same time, when q and p change, and $\alpha$ and $\beta$ are both 0.5, the result of the score function of the scheme is shown in Figure 5. It can be seen from the results in Figure 5 that when p and q change at the same time, it will not affect the ordering of the schemes.

Table 18 The ranking of schemes under different p and q

| Scheme ranking | p=2 | p=3 | p=4 | p=5 |
| --- | --- | --- | --- | --- |
| q=3 | $y_2>y_1>y_3>y_4>y_5$ | $y_2>y_1>y_3>y_4>y_5$ | $y_2>y_1>y_3>y_4>y_5$ | $y_2>y_1>y_3>y_4>y_5$ |
| q=4 | $y_2>y_1>y_3>y_4>y_5$ | $y_2>y_1>y_3>y_4>y_5$ | $y_2>y_1>y_3>y_4>y_5$ | $y_2>y_1>y_3>y_4>y_5$ |
| q=5 | $y_2>y_1>y_3>y_4>y_5$ | $y_2>y_1>y_3>y_4>y_5$ | $y_2>y_1>y_3>y_4>y_5$ | $y_2>y_1>y_3>y_4>y_5$ |
| q=6 | $y_2>y_1>y_3>y_4>y_5$ | $y_2>y_1>y_3>y_4>y_5$ | $y_2>y_1>y_3>y_4>y_5$ | $y_2>y_1>y_3>y_4>y_5$ |
| q=7 | $y_2>y_1>y_3>y_4>y_5$ | $y_2>y_1>y_3>y_4>y_5$ | $y_2>y_1>y_3>y_4>y_5$ | $y_2>y_1>y_3>y_4>y_5$ |

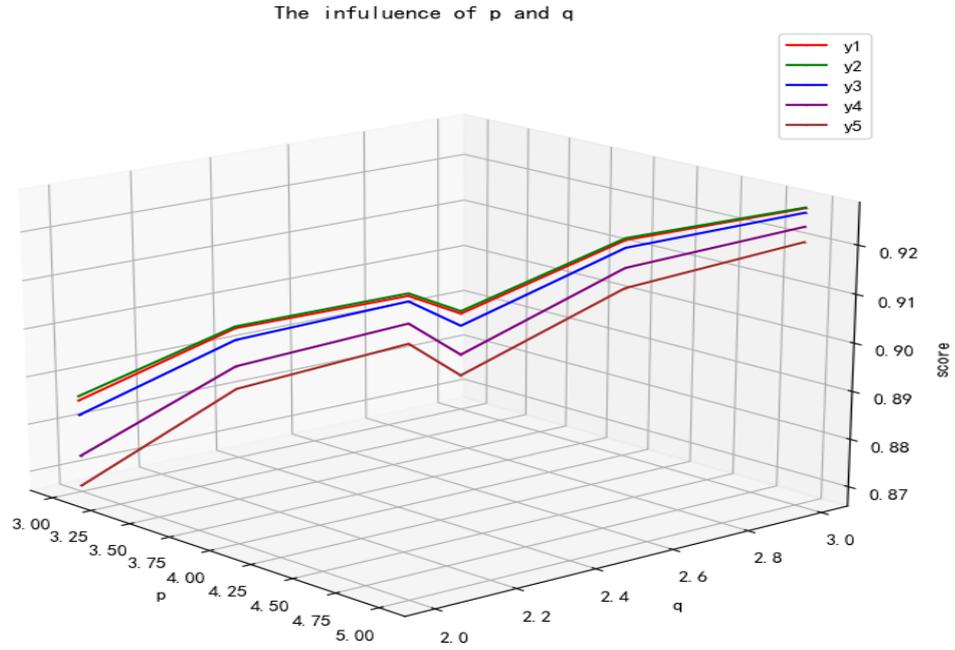

Figure 5 Sorting results of schemes under different p and q

When studying the impact of changes in p and q on the ranking of options in the previous article, the values of $\alpha$ $and$ $\beta$ in the score function were fixed. In order to ensure the effectiveness of the group decision-making method proposed in this paper, the impact of changes in sum on the ranking of options was further explored. Therefore, When p=2, q=3 and $\alpha(0<\alpha<1)$ randomly changes, the result of the score function of the scheme is shown in Figure 6. It can be seen from the results in Figure 6 that when $\alpha$ changes, it will not affect the ordering of the schemes. Because of $\beta=1-\alpha$, $\beta$ will not affect the ordering of the scheme.

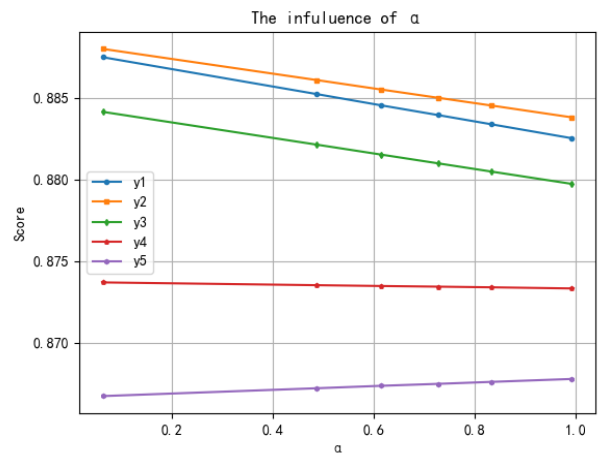

Figure 6 Sorting results of schemes under different $\alpha$

### 7.2.2 Attribute weight analysis

After studying the influence of the parameters on the scheme ranking, I began to compare and analyze the influence of different distance measurement methods on the attribute weight and expert weight determination in the scheme ranking. Distance measurement methods can be divided into three types: positive ideal distance, negative ideal distance and centered distance. The reference point for the positive ideal distance is $<[1,1],[0,0]>$, the reference point for the negative ideal distance is $<[0,0],[1,1]>$, and the reference points for the middle distance are $<[1,1],[0,0]>$ and $<[0,0],[1,1]>$. The formulas for calculating attribute weights and expert weights in the middle distance are consistent with the steps used in this article to calculate attribute weights and expert weights with negative ideal distances, while the steps to calculate attribute weights and expert weights with positive ideal distances are to calculate attribute weights at negative ideal distances. Use formula (68) to normalize $1-\omega_j (j=1,2,...,n)$ based on the steps of expert weight and expert weight:

Table 19 Weights measured by different distance methods

| Distance method | Attribute real number weight (retain 5 decimal places) |
|---|---|
| Positive ideal distance | $\omega_1=0.20100$ |
| | $\omega_2=0.19978$ |
| | $\omega_3=0.20029$ |
| | $\omega_4=0.19957$ |
| | $\omega_5=0.19937$ |
| Negative ideal distance | $\omega_1=0.20384$ |
| | $\omega_2=0.19914$ |
| | $\omega_3=0.20110$ |
| | $\omega_4=0.19835$ |
| | $\omega_5=0.19757$ |
| Middle distance | $\omega^*_1=0.2$ |
| | $\omega^*_2=0.2$ |
| | $\omega^*_3=0.2$ |
| | $\omega^*_4=0.2$ |
| | $\omega^*_5=0.2$ |

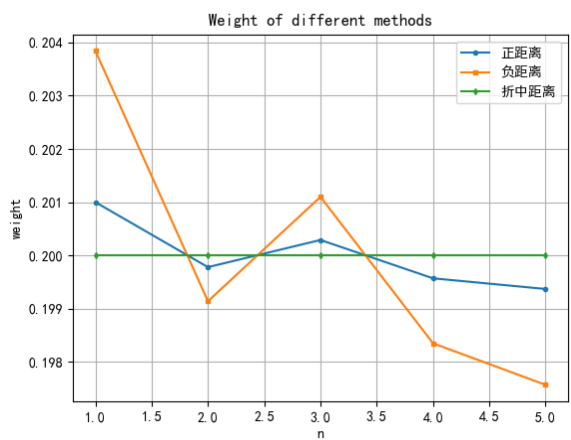

Figure 7 Attribute weights under different methods

In order to make a more scientific comparison and analysis, the order of the schemes calculated by the group decision-making method through the ideal distance is shown in Table 20. From the results in Table 20, it can be found that the calculation results of different ideal distances are all orange warnings. Then, by observing Figure 7 we can find that, compared to the case of this article, when the calculation results of different ideal distances are the same, the volatility of the calculation results of the negative ideal distance is greater than the calculation results of the positive ideal distance and the middle distance. Reflect the importance of the attributes, therefore, the negative ideal distance has more advantages in this case.

Table 20 The ranking of the schemes measured by different ideal distances

| Ideal distance | Score sort | Scheme ranking | decision making |
|---|---|---|---|
| Positive ideal distance | $Sr_2>Sr_1>Sr_3>Sr_4>Sr_5$ | $y_2>y_1>y_3>y_4>y_5$ | Orange warning |
| Negative ideal distance | $Sr_2>Sr_1>Sr_3>Sr_4>Sr_5$ | $y_2>y_1>y_3>y_4>y_5$ | Orange warning |
| Middle distance | $Sr_2>Sr_1>Sr_3>Sr_4>Sr_5$ | $y_2>y_1>y_3>y_4>y_5$ | Orange warning |

By comparing and analyzing the parameters included in the group decision-making method proposed in this article in section 7.2, it can be found that the changes in the parameters of the entire group decision-making method will not affect the ranking changes of the schemes. Therefore, in section 7.2, the parameters proposed in this article can be determined The group decision-making method is effective.

### 7.2.3 Comparative analysis of operators

In order to further verify the effectiveness and feasibility of the group decision-making method proposed in this paper, this paper will use the five operators in the literature [41][32][23] to replace the IVq-ROFYWA operator in the calculation process of the 7.1 case. Calculate, finally compare their calculation results with the calculation results of the operators proposed in this article. Regarding the parameters of the operators, all operators have q=3, $x = 2$ in q-RIVOFWHM, $k = 2$ in IVq-ROFWMSM, $\theta = 2$ in q_RIVOFFWA and q_RIVOFFGA, and the calculation results of 5 operators are as follows Table 21 shows.

Table 21 Comparison results under different operators

| Operator name | Score sort | Scheme ranking | decision making |
|---|---|---|---|
| q-RIVOFWHM[41] | $Sr_2>Sr_1>Sr_3>Sr_4>Sr_5$ | $y_2>y_1>y_3>y_4>y_5$ | Orange warning |
| IVq-ROFWMSM[32] | $Sr_2>Sr_1>Sr_3>Sr_4>Sr_5$ | $y_2>y_1>y_3>y_4>y_5$ | Orange warning |
| q-RIVOFWA[41] | $Sr_2>Sr_1>Sr_3>Sr_4>Sr_5$ | $y_2>y_1>y_3>y_4>y_5$ | Orange warning |
| q-RIVOFFWA[23] | $Sr_2>Sr_1>Sr_3>Sr_4>Sr_5$ | $y_2>y_1>y_3>y_4>y_5$ | Orange warning |
| q-RIVOFFGA[23] | $Sr_2>Sr_1>Sr_3>Sr_4>Sr_5$ | $y_2>y_1>y_3>y_4>y_5$ | Orange warning |
| **IVq-ROFYWA** | **$Sr_2>Sr_1>Sr_3>Sr_4>Sr_5$** | **$y_2>y_1>y_3>y_4>y_5$** | **Orange warning** |

Results in Table 16 indicate that the IVq-ROFYWA operator proposed in this paper is in the same order as the operator in the literature [41][32][23]. Finally, the resident status is shown as an orange warning, and it is at the same time as 5 The early warning schemes given by two hypertension experts are consistent, thus verifying the effectiveness of the operator proposed in this paper and the feasibility of the group decision method proposed in this paper.

### 7.2.4 Comparative analysis of decision-making methods

Section 7.3 of this paper compares the operator proposed in this paper with other operators in the group decision making method, and observes that different operators combine with the decision method proposed in this paper to the decision result of the case in 7.1 of this paper. Because the results are consistent, it verifies this paper. The feasibility of the proposed decision-making method in the group decision-making method. In order to further verify the effectiveness and feasibility of the group decision-making method proposed in this article, this article will use the decision-making methods of [18][27][45][17][32] to replace the WASPAS method in the calculation process of the 7.1 case. , And finally compare their decision-making results with the calculation results of the operators proposed in this paper. The calculation results of different decision-making methods are shown in Table 22.

Table 22 Comparison results under different decision-making methods

| Method | Score sort | Alternative ranking | Result |
|---|---|---|---|
| MOORA[18] | $Sr_2>Sr_1>Sr_3>Sr_4>Sr_5$ | $y_2>y_1>y_3>y_4>y_5$ | Orange warning |
| COPRAS[27] | $Sr_5>Sr_2>Sr_1>Sr_4>Sr_3$ | $y_2>y_1>y_3>y_4>y_5$ | Orange warning |
| MACBETH[45] | $Sr_2>Sr_1>Sr_3>Sr_4>Sr_5$ | $y_2>y_1>y_3>y_4>y_5$ | Orange warning |
| MABAC[17] | $Sr_2>Sr_1>Sr_3>Sr_4>Sr_5$ | $y_2>y_1>y_3>y_4>y_5$ | Orange warning |
| IVq-ROF-ARAS[32] | $Sr_2>Sr_1>Sr_3>Sr_4>Sr_5$ | $y_2>y_1>y_3>y_4>y_5$ | Orange warning |
| **WASPAS** | $\mathbf{Sr_2>Sr_1>Sr_3>Sr_4>Sr_5}$ | $\mathbf{y_2>y_1>y_3>y_4>y_5}$ | **Orange warning** |

According to the results in Table 18, it can be found that the decision-making method used in this article is different from the COPRAS method, indicating that the resident's status is a green warning. In-depth study of the reasons for the different ranking results of the COPRAS method, according to the characteristics of the COPRAS method, the key formula is as shown in (69).

$$Q_i = S_{+i} + \frac{\sum_{i=1}^{m} S_{-i}}{S_{-i}\sum_{i=1}^{m} \frac{1}{S_{-i}}} \qquad (69)$$

The COPRAS method uses formula (69) to determine the relative saliency value $Q_i$(i=1,2,…,n) of each scheme and then determines the order of the schemes according to the relative saliency value, where $S_{+i}$ is the minimum positive attribute of the i-th row Index, $S_{-i}$ is the minimum index of the negative attribute of the i-th row. Therefore, the COPRAS method is suitable for solving the multi-attribute decision-making problem where the benefit attribute and the cost attribute exist at the same time, but the case presented in this paper does not satisfy this condition , so the ranking results will be different. However, the difference in the ranking of one method does not affect the overall comparison. The decision results of other methods finally show that the resident's status is an orange warning, which is consistent with the warning plan given by 5 hypertension experts.

Through Section 5.3, we know that the WASPAS method is based on the fusion of WSM and WPM preferences, which increases the accuracy of the data. The advantage of the WSM model is that it is more sensitive and can reflect differences quickly, but the disadvantage is that it is more susceptible to extreme numbers. The advantage of the WPM model is that it is not easily affected by extreme numbers and is very stable. The disadvantage is that it cannot be reflected in time when it is necessary to reflect the differences. WASPAS cleverly integrates the two models through the preference of the two, and the data reflected is more accurate. In order to verify the characteristics of the WASPAS method, this article will use WSM and WPM to replace the WASPAS method in the proposed group decision making calculations. At this time, $\lambda = 0.5$，compare and analyze their ranking results and score values. The calculation results are shown in Table 23 and Figure 8. Shown.

Table 23 Comparison results under different models

| Method | Score sort | Rank |
|---|---|---|
| WSM | $Sr_2(0.88820)>Sr_1(0.88718)>Sr_3(0.88436)>Sr_4(0.87623)>Sr_5(0.87041)$ | $y_2>y_1>y_3>y_4>y_5$ |
| WPM | $Sr_2(0.88380)>Sr_1(0.88307)>Sr_3(0.87968)>Sr_4(0.87065)>Sr_5(0.86382)$ | $y_2>y_1>y_3>y_4>y_5$ |
| **WASPAS** | $\mathbf{Sr_2(0.88604)>Sr_1(0.88517)>Sr_3(0.88207)>Sr_4(0.87352)>Sr_5(0.86722)}$ | $\mathbf{y_2>y_1>y_3>y_4>y_5}$ |

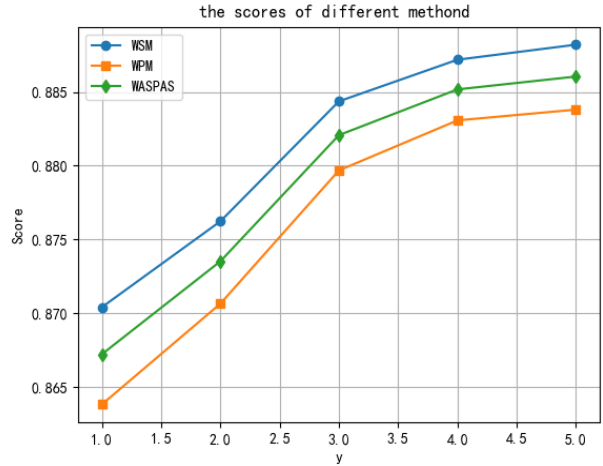

Figure 8 Attribute weights under different models

From the results in Table 23 and Figure 8, it can be found that the calculation results of the WSM model and WPM model for this case are consistent with the calculation results of WASPAS. When $\lambda = 0.5$, the calculation result of WASPAS is in the middle of the calculation result of WSM and WPM; when $\lambda > 0.5$, the result of WASPAS is biased toward WSM, and the WASPAS method is more sensitive at this time; when $\lambda < 0.5$, the result of WASPAS is biased toward WPM, at this time The WASPAS method is relatively stable. In addition, the calculation result of WASPAS always fluctuates between the calculation result of WSM and the calculation result of WPM, and the order of the scheme remains unchanged. It further verifies the more accuracy of WASPAS decision-making results and the feasibility of the group decision-making method proposed in this paper.

## 8. Conclusion

The interval-valued group decision-making method under the generalized orthogonal fuzzy environment with unknown weights described in this paper integrates the proposed IVq-ROFYWA, IVq-ROFYWG operators, and the CRITIC-WASPAS decision-making method, which provides people with a new decision-making tool. The group decision-making method proposed in this paper is carried out in an interval-valued generalized orthogonal fuzzy environment, so decision makers have more freedom when making fuzzy judgments of uncertainties. The integrated group decision-making method not only considers the uncertainties of different experts. It also takes into account the impact of variability and relevance on indicators. The group decision-making method has the following characteristics: (1) Eliminate the uncertainty of expert evaluation and obtain intensive data information; (2) Eliminate the influence of some relatively strong indicators, reduce the overlap of information between indicators, and obtain (3) The importance of attributes can be analyzed based on expert decision data, that is, the ranking of attribute weights; (4) The evaluation value of the scheme is highly accurate. Specific contributions include:

This study expanded the interval-valued generalized orthogonal fuzzy distance measurement method, and proposed a new expert solution weight solution method based on the ideal distance. While solving the expert weight, it avoids the decision maker from affecting other solutions due to external influences. the weight of. This paper proposes a new scoring function to compare the size of IVq-ROFNs and compare and analyze with the existing scoring function. The analysis results show that the new scoring function successfully overcomes the shortcomings of the existing scoring function. Based on the fuzzy Yager operator and algorithm, IVq-ROFYWA and IVq-ROFYWG operators are developed and integrated with the decision-making method proposed in this paper. The CRITIC and WASPAS decision-making methods are extensively studied, and they are extended to the interval-valued generalized orthogonal fuzzy environment, and combined with the IVq-ROFYWA and IVq-ROFYWG operators to obtain the CRITIC-WASPAS decision method, which can be used to solve the MADM problem. Based on the integration of IVq-ROFYWA operator, decision maker weight, CRITIC-WASPAS decision method, and scoring function, a new interval-value generalized orthogonal group decision-making method is proposed. At the end of this article, the proposed group decision-making is applied to the application of hypertension risk management, and the group decision-making proposed in this article is compared and analyzed from the three aspects of parameter change, operator change and method change. The experimental and analysis results show that: (1) Case The result of the calculation is in the parameters p, q,$\alpha$、 $\beta$ are changed;(2) When the calculation results of different ideal distance cases are the same, the negative ideal distance has greater volatility and can better reflect the difference; (3) The case calculation results are the same as the operator calculation results that have been proposed by experts; (4) The calculation results of the case are the same as those of other decision-making methods.

The group decision-making research in this paper is based on the decision matrix. Another current research hotspot is how to solve the problem of consistency and consensus among experts after the decision makers give a judgment matrix. In the future, we will combine the consistency problem of the judgment matrix and the consensus problem to study the weights. Unknown group decision making method [72]. Although the group decision-making method proposed in this paper can solve the decision-making environment with high data information density, it will continue to analyze and study whether it is suitable for other decision-making environments. At the same time, the decision-making method cases presented in this paper are based on a small amount of decision-maker data, but the actual application requires a large amount of data verification and analysis. Therefore, the future will further integrate big data [73],Artificial intelligence to deal with complex intelligent decision-making problems with unknown weights [74].